%% file: main.tex
\documentclass[11pt, logo, onecolumn, copyright, colorlinks=true, allcolors=blue]{nvidiatechreport}
\usepackage{graphicx} 
\usepackage{subcaption,ragged2e}
\usepackage[round]{natbib}

\usepackage[normalem]{ulem}
\usepackage[utf8]{inputenc} 
\usepackage[T1]{fontenc}    
\usepackage{url}
\usepackage{tikz}
\usepackage{enumitem}
\usetikzlibrary{positioning}
\usetikzlibrary{calc}
\usepackage{array}
\usepackage{booktabs}
\usepackage{wrapfig}
\usepackage{caption} 
\usepackage{listings}
\usepackage{adjustbox}
\usepackage{footnote}
\usepackage{multirow}
\usepackage{amsmath}
\usepackage{amsfonts}
\usepackage{breakcites}
\usepackage{blindtext}
\usepackage{svg}
\usepackage[most]{tcolorbox}
\usepackage{tabularx}
\usepackage{graphicx} 
\usepackage{xcolor}
\usepackage{float}
\usepackage{needspace}
\usepackage{makecell}
\usepackage{colortbl}
\newcommand{\newparagraph}[1]{\noindent\textbf{#1\hspace{0.5em}}}

\newcommand{\ourmodelfull}{NVIDIA Nemotron 3 Super\xspace}
\newcommand{\ourmodel}{Nemotron 3 Super\xspace}
\newcommand{\ourbasemodel}{Nemotron 3 Super 120B-A12B Base\xspace}

\newcommand{\nanomodel}{Nemotron 3 Nano\xspace}

\title{\ourmodel: Open, Efficient Mixture-of-Experts Hybrid Mamba-Transformer Model for Agentic Reasoning}
\author{\large NVIDIA}
\date{}

\begin{document}

\begin{abstract}
\large \textbf{Abstract.}
\input{sections/abstract.tex}
\end{abstract}

\maketitle

\input{sections/introduction}
\input{sections/pretraining}

\clearpage
\input{sections/alignment}

\input{sections/quantization}
\input{sections/conclusion}
\input{sections/contributors}

\bibliography{references}
\bibliographystyle{references}

\input{sections/appendix}

\end{document}

%% file: sections/abstract.tex
\normalsize


We describe the pre-training, post-training, and quantization of \ourmodel, a 120 billion (active 12 billion) parameter hybrid Mamba-Attention Mixture-of-Experts model. Nemotron 3 Super is the first model in the Nemotron 3 family to 1) be pre-trained in NVFP4, 2) leverage LatentMoE, a new Mixture-of-Experts architecture that optimizes for both accuracy per FLOP and accuracy per parameter, and 3) include MTP layers for inference acceleration through native speculative decoding. We pre-trained Nemotron 3 Super on 25 trillion tokens followed by post-training using supervised fine tuning (SFT) and reinforcement learning (RL). The final model supports up to 1M context length and achieves comparable accuracy on common benchmarks, while also achieving up to 2.2$\times$ and 7.5$\times$ higher inference throughput compared to GPT-OSS-120B and Qwen3.5-122B, respectively. Nemotron 3 Super datasets, along with the base, post-trained, and quantized checkpoints, are open-sourced on HuggingFace.

%% file: sections/introduction.tex
\section{Introduction}
\label{sec:intro}


The last few years have seen a rise in the popularity of Mixture-of-Experts (MoE) based Large Language Models (LLMs)~\citep{deepseekai2025deepseekv3technicalreport,yang2025qwen3technicalreport,5team2025glm45agenticreasoningcoding}. MoEs help LLMs achieve higher accuracy at a lower active parameter count than regular dense models~\citep{dai2024deepseekmoe,lepikhin2020gshard}. Orthogonal to MoEs, Hybrid Mamba-Attention models have shown promise in significantly improving inference throughput~\citep{nemotronnanov2}. We combine these two directions of improvement in Nemotron 3~\citep{nvidia2025nvidianemotron3efficient}. As part of our Nemotron 3 series of models, we present \ourmodel---a 12 billion active, 120 billion total parameter MoE hybrid Mamba-Attention model. \ourmodel achieves better or on-par benchmark accuracies than GPT-OSS-120B~\citep{openai2025gptoss120bgptoss20bmodel} and Qwen3.5-122B while achieving up to 2.2$\times$ and 7.5$\times$ higher inference throughput, respectively, on the 8k token input / 64k token output setting.   

\begin{figure}[ht]
    \centering
    \includegraphics[width=\linewidth]{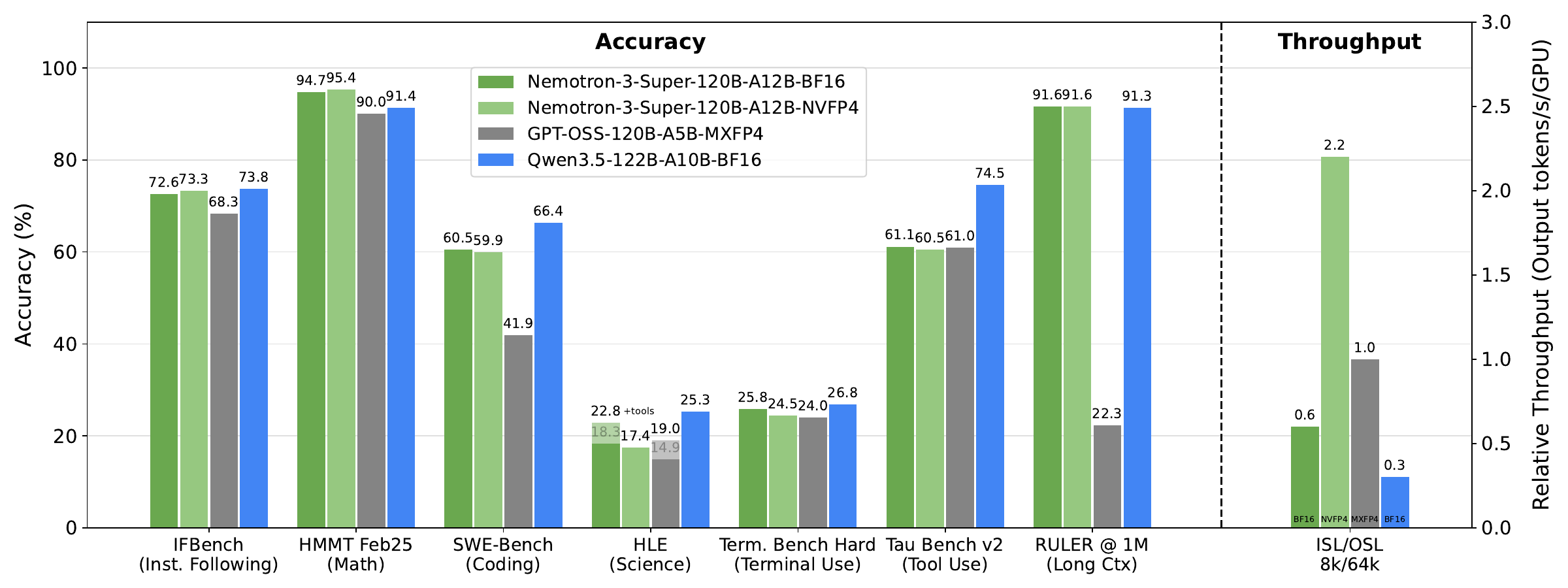}
    \caption{Accuracy and throughput comparisons of \ourmodel with GPT-OSS-120B and Qwen3.5-122B. \ourmodel achieves comparable accuracies across popular benchmarks but provides the highest inference throughput; on 8k input sequence lengths and 64k output sequence lengths, \ourmodel provides up to 2.2$\times$ and 7.5$\times$ higher throughput than GPT-OSS-120B and Qwen3.5-122B, respectively. We measured throughput on B200 GPUs with vLLM and TRT-LLM and use the best out of the two frameworks for each model. For GPT-OSS-120B we use MXFP4 weights, MXFP8 activations, and FP8 KV-Cache; for Qwen3.5-122B we use BF16. We used the OpenHands harness to evaluate SWE-Bench.}
    \label{fig:intro}
\end{figure}

\ourmodel is our first model to use LatentMoE~\citep{latentmoe_tr} - a novel MoE architecture that achieves better accuracy per parameter and per FLOP than regular MoEs. \ourmodel also incorporates Multi-Token-Prediction (MTP), which accelerates inference through speculative decoding while improving overall model quality. We pre-trained \ourmodel in NVFP4, demonstrating stable and accurate pre-training in low precision. Similar to Nemotron 3 Nano~\citep{nvidia2025nemotron3nanoopen}, we pre-trained \ourmodel on 25 trillion text tokens divided into 2 phases. The first phase accounted for 80\% of pre-training (20 trillion tokens) and focused on diversity and broad coverage, while the second phase accounted for 20\% of pre-training (5 trillion tokens) and focused on high-quality data and benchmark accuracy. Our base model achieves significantly better accuracy than similarly sized state-of-the-art base models, such as GLM-4.5-Air-Base~\citep{5team2025glm45agenticreasoningcoding} and Ling-flash-Base-2.0~\citep{lingteam2025every}. 

We trained \ourmodel with a strong emphasis on agentic capabilities. To support this objective, we substantially scaled the breadth of our RL environments, the volume and quality of agentic training data, and the overall amount of post-training focused on multi-step tool-using behavior. To train effectively on this diverse set of long-horizon tasks, we made substantial improvements to the resiliency of our RL infrastructure, enabling large-scale asynchronous training. This expanded agentic training recipe yields substantial improvements over Nemotron 3 Nano across software engineering, terminal use, and general tool use benchmarks.

We are publicly sharing the training recipe for \ourmodel on the Nemotron Developer Repository\footnote{\href{https://github.com/NVIDIA-NeMo/Nemotron}{https://github.com/NVIDIA-NeMo/Nemotron}}. We are also openly releasing the following:

\newparagraph{Checkpoints}
\begin{itemize}
    \item \href{https://huggingface.co/nvidia/NVIDIA-Nemotron-3-Super-120B-A12B-NVFP4}{\texttt{Nemotron 3 Super 120B-A12B NVFP4} \includegraphics[height=0.9em]{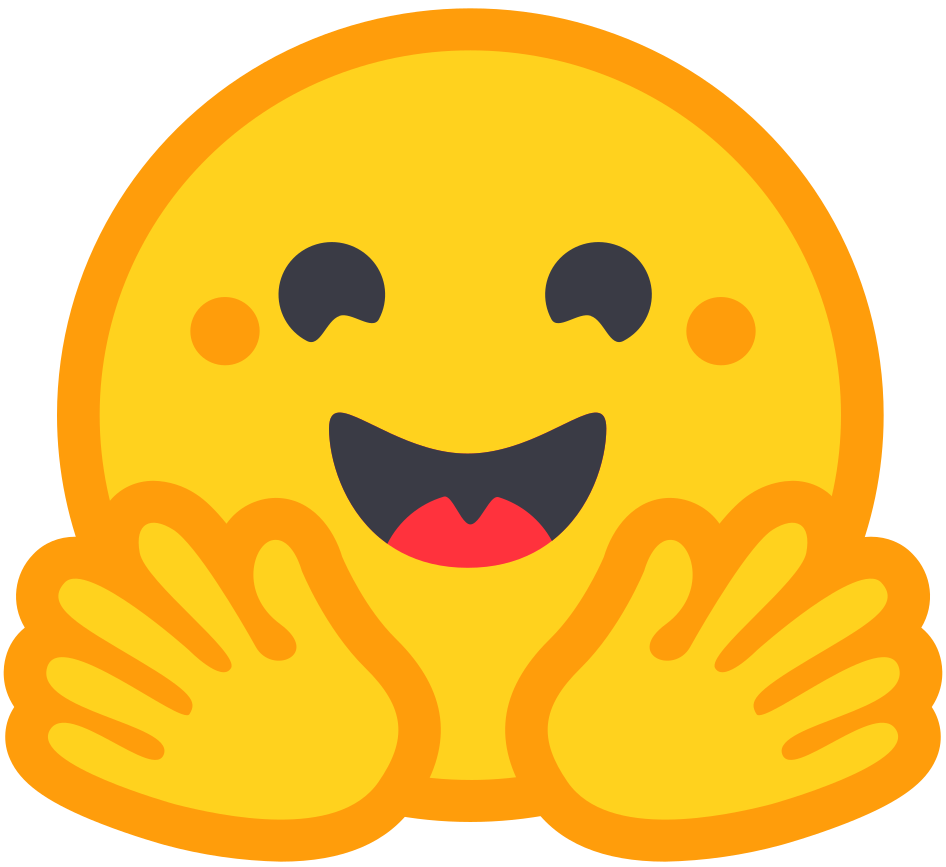}} : post-trained and NVFP4 quantized model
    \item \href{https://huggingface.co/nvidia/NVIDIA-Nemotron-3-Super-120B-A12B-FP8}{\texttt{Nemotron 3 Super 120B-A12B FP8} \includegraphics[height=0.9em]{assets/huggingface-color.png}} : post-trained and FP8 quantized model
    \item \href{https://huggingface.co/nvidia/NVIDIA-Nemotron-3-Super-120B-A12B-BF16}{\texttt{Nemotron 3 Super 120B-A12B BF16} \includegraphics[height=0.9em]{assets/huggingface-color.png}} : post-trained model
    \item \href{https://huggingface.co/nvidia/NVIDIA-Nemotron-3-Super-120B-A12B-Base-BF16}{\texttt{Nemotron 3 Super 120B-A12B Base BF16} \includegraphics[height=0.9em]{assets/huggingface-color.png}} : base model
    \item \href{https://huggingface.co/nvidia/Qwen3-Nemotron-235B-A22B-GenRM-2603}{\texttt{Qwen3-Nemotron-235B-A22B-GenRM-2603} \includegraphics[height=0.9em]{assets/huggingface-color.png}} : GenRM used for RLHF
\end{itemize}

\newparagraph{Data}
\begin{itemize}
    \item \href{https://huggingface.co/datasets/nvidia/Nemotron-Pretraining-Specialized-v1.1}{\texttt{Nemotron-Pretraining-Specialized-v1.1} \includegraphics[height=0.9em]{assets/huggingface-color.png}} : a collection of synthetic datasets aimed to improve LLM capabilities in code concepts and algorithms, formal logic, economics, and multiple choice questions.
    \item \href{https://huggingface.co/collections/nvidia/nemotron-post-training-v3}{\texttt{Nemotron-Super-Post-Training-Data} } \includegraphics[height=0.9em]{assets/huggingface-color.png} : a collection of RL environments and SFT datasets targeting a broad range of agentic capabilities.
\end{itemize}

The report is organized into 3 broad sections: Pre-training~(\S\ref{sec:pretraining}), Post-training~(\S\ref{sec:alignment}), and Quantization~(\S\ref{sec:quantization}), each describing in detail our approach.

%% file: sections/pretraining.tex
\section{Pretraining}
\label{sec:pretraining}


In this section, we highlight the key features of \ourbasemodel, detailing its hybrid Mamba-Attention Mixture-of-Experts (MoE) architecture, NVFP4 pre-training, hyperparameter configurations, long-context extension, and the 25-trillion-token corpus used for pretraining. We also demonstrate that Nemotron-3 Super 120B A12B Base achieves superior accuracy compared to other public state-of-the-art models—including Ling-flash-Base-2.0 and GLM-4.5-Air-Base—across a comprehensive suite of benchmarks. 

\input{sections/pretraining/architecture}

\input{sections/pretraining/nvfp4}
\input{sections/pretraining/data}

\input{sections/pretraining/training}

\input{sections/pretraining/base_model_evals}

%% file: sections/pretraining/architecture.tex
\subsection{Model Architecture}
\label{sec:arch}

\begin{figure}[!h]
\centering
\includegraphics[width=0.9\linewidth]{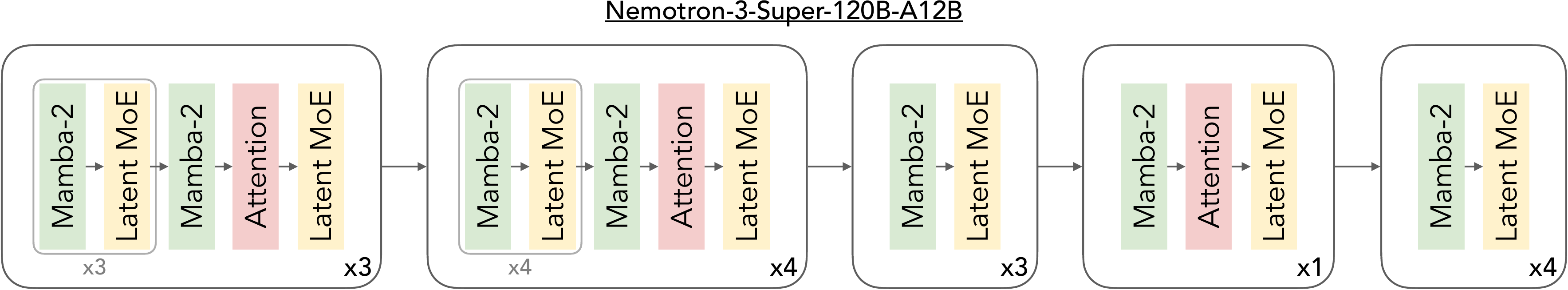}
\caption{\ourmodel layer pattern. Similar to \nanomodel, we use a hybrid Mamba-Attention architecture, but \ourmodel is the first model to scale sparsely using LatentMoE layers rather than standard MoE layers.}
\label{fig:layer-pattern}
\end{figure}

\begin{table*}[!hbt]\small\centering
\renewcommand{\arraystretch}{1.2}
\setlength{\tabcolsep}{10pt}
\begin{tabular}{l|c}
\toprule
\textbf{Configuration} & \textbf{\ourbasemodel} \\
\midrule
Total Layers & 88 \\
Model Dimension  & 4096 \\
\midrule
Q-Heads ($n_{q}$) & 32 \\
KV-Heads ($n_{kv}$) & 2 \\
Head Dimension & 128 \\
\midrule
Mamba State Dimension & 128 \\
Mamba Groups & 8  \\ 
Mamba Heads & 128 \\
Mamba Head Dimension & 64 \\ 
\midrule
Expert Hidden Dimension & 2688 \\
Shared Expert Intermediate Size & 5376 \\
Total Experts per Layer & 512 \\
Top-$k$ (Activated Experts) & 22 \\
\midrule
MoE Latent Size & 1024 \\
\midrule
MTP layers (shared weight) & 2 \\ 
\bottomrule
\end{tabular}
\caption{\ourmodel Architectural Dimensions. The model employs a hybrid Mamba-2 and MoE design with strategic global attention layers to optimize the balance between sequence modeling performance and inference throughput.}
\label{table:model_arch}
\end{table*}

\ourbasemodel scales up the hybrid Mamba-Attention Mixture-of-Experts (MoE) architecture introduced in Nemotron-3 Nano~\citep{nvidia2025nvidianemotron3efficient}. We extend this foundation to 120.6B total parameters, maintaining a constrained active budget of 12.7B parameters (12.1B excluding embeddings) per forward pass. The architecture comprises three core pillars: sparse LatentMoE scaling~(\S\ref{sec:latent_moe}), Multi-Token Prediction (MTP) for inference acceleration~(\S\ref{sec:mtp}), and a periodic hybrid interleaving pattern~(\S\ref{sec:hybrid_moe_arch}).




\subsubsection{LatentMoE: Hardware-Aware Expert Design for Improved Accuracy per Byte} \label{sec:latent_moe}
Mixture-of-Experts (MoE) architectures have emerged as a promising approach to maximize accuracy under fixed inference cost, allowing models to scale in parameter count while keeping floating-point operations (FLOPs) per token constant.
Existing MoE designs are largely motivated by high-level sparsity arguments and optimized for offline, throughput-oriented settings, with little consideration for online deployments that impose strict latency, memory bandwidth, and communication constraints.
Whereas accuracy per FLOP reflects computational efficiency, accuracy per parameter captures memory footprint, memory bandwidth, routing-induced communication, and sharding overhead.
Neglecting these factors can yield architectures that appear efficient in aggregate compute yet incur substantial inefficiency in practice.

Motivated by these observations, we revisited MoE design from a hardware--software co-design perspective.
Through systematic analysis of existing MoE systems across the throughput--latency Pareto frontier, together with accuracy measurements and theoretical analysis, we identified structural inefficiencies in prevailing MoE designs that limit accuracy per unit inference cost.
From this analysis we distilled the following design principles for efficient MoE scaling:
\begin{enumerate}
    \setlength{\itemsep}{0.4em}
    \item In low-latency serving, MoE inference is often dominated by the memory bandwidth cost of reading expert weights. Each expert matrix has size $d \times m$, where $d$ is the hidden dimension and $m$ is the expert FFN intermediate dimension; reducing this cost therefore requires decreasing $d$ or $m$.
    \item In throughput-oriented serving, distributed MoE inference is dominated by all-to-all routing. Routing volume scales as $d \times K$, where $K$ is the number of active experts; reducing communication overhead therefore requires decreasing $d$ or $K$.
    \item Preserving model quality requires preserving the effective nonlinear budget $K \cdot m$. To relieve memory and communication bottlenecks without sacrificing quality, $K$ and $m$ should thus be held fixed.
    \item A task-specific effective feature rank $r_{\mathrm{eff}}$ imposes a lower limit on how much $d$ can be reduced; reducing $d$ below this limit causes model quality to collapse.
    \item Scaling both the total number of experts $N$ and the top-$K$ experts per token improves quality by exponentially expanding the space of expert combinations.
\end{enumerate}
Principles (1)--(3) imply that the hidden dimension $d$ is the most promising axis for reduction, enabling gains in both throughput- and latency-oriented regimes without significant loss in accuracy.
Principle (4) gives a lower bound on how far $d$ can be reduced without collapse.
Principle (5) indicates that increasing $N$ and $K$ improves quality; because memory bandwidth and communication scale linearly with $K$, we can increase $K$ by a factor $\alpha$ and reduce $d$ by the same factor $\alpha$ to obtain higher accuracy at similar inference cost.
Guided by these insights, we developed LatentMoE~\citep{latentmoe_tr}, a MoE architecture designed to achieve higher accuracy than a standard MoE at similar inference cost.

\begin{figure}[h!]
    \centering 
    \begin{subfigure}[t]{0.49\textwidth}
        \centering
        \includegraphics[width=\linewidth]{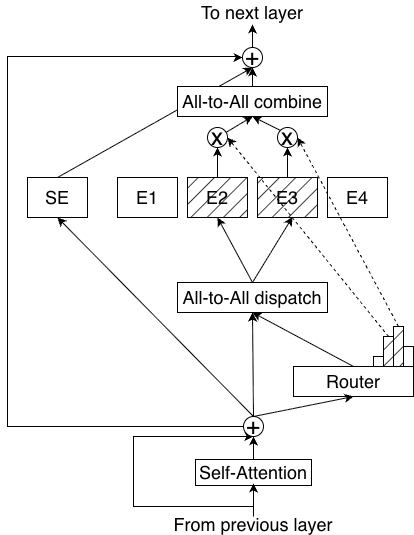}
        \caption{Standard MoE architecture.}
        \label{fig:standard_moe}
    \end{subfigure}
    \hfill
    \begin{subfigure}[t]{0.49\textwidth}
        \centering
        \includegraphics[width=\linewidth]{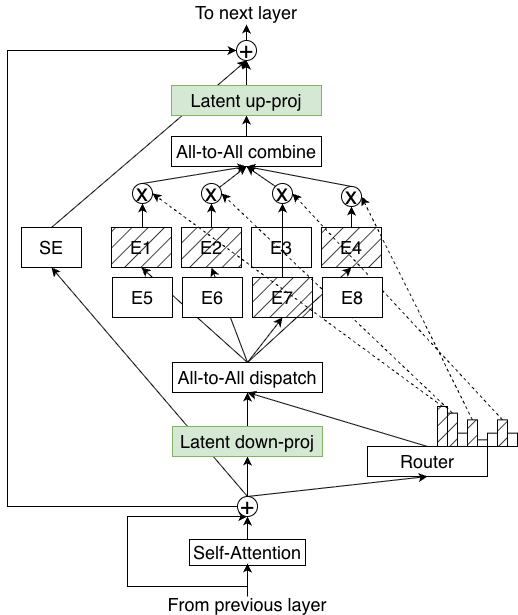}
        \caption{LatentMoE architecture.}
        \label{fig:latent_moe}
    \end{subfigure}
     \caption{Standard MoE vs.\ LatentMoE. In LatentMoE, tokens are projected from the hidden dimension $d$ into a smaller latent dimension $\ell$ for routing and expert computation, reducing routed parameter loads and all-to-all traffic by a factor $d/\ell$. These savings are used to increase both the total number of experts and the top-$K$ active experts per token by the same factor, improving model accuracy at approximately constant inference cost.}
    \label{fig:intro_fig}
\end{figure}


The LatentMoE architecture is illustrated in Figure~\ref{fig:latent_moe}.
Each input token $x \in \mathbb{R}^d$ is first projected into a lower-dimensional latent space $\mathbb{R}^{\ell}$ via a learnable down-projection matrix $W_{\downarrow} \in \mathbb{R}^{\ell \times d}$.
The compressed representation is then routed to an expanded set of experts that operate entirely in this latent space.
After expert computation, the outputs are aggregated and projected back to dimension $d$ via a learnable up-projection matrix $W_{\uparrow} \in \mathbb{R}^{d \times \ell}$.
Shifting routed expert computation and all-to-all traffic into the latent space reduces both per-expert weight loads and communication payloads by a factor $d/\ell$ relative to a standard MoE.
We use these savings to increase the total number of experts from $N$ to $N' = N \cdot d/\ell$ and the top-$K$ active experts per token from $K$ to $K' = K \cdot d/\ell$.
The reduction in dimension offsets the increase in expert count and in $K$, yielding higher model quality at a similar computational and communication budget.
To preserve quality, all non-routed computations---including the routing gate (gating network), shared expert computation, and non-expert layers---remain in the full hidden dimension $d$, as they do not contribute significantly to the targeted bottlenecks.
We refer the reader to the LatentMoE technical report~\citep{latentmoe_tr} for further details.

\subsubsection{Multi-Token Prediction} \label{sec:mtp}
Nemotron-3 Super incorporates a Multi-Token Prediction (MTP) objective to improve both modeling quality and inference efficiency. Unlike conventional next-token training, MTP optimizes the model to predict multiple future tokens at each position~\citep{gloeckle2024better, deepseekai2025deepseekv3technicalreport}. This encourages representations that capture multi-step dependencies and longer-range structure, leading to consistent improvements in validation loss and downstream benchmark accuracy.

Beyond quality gains, MTP enables native speculative decoding. The auxiliary prediction heads function as an internal draft model: during inference, they generate candidate continuations that are verified by the main model in a single forward pass. This substantially reduces decoding latency while introducing minimal additional FLOPs—significantly less than required by an external draft model. Although speculative decoding is particularly effective at small batch sizes, recent work shows that it can also improve throughput in larger-batch and sparse MoE settings~\citep{huang2025moesd}.

\paragraph{Design for Robust Autoregressive Drafting.} Standard MTP implementations use $N$ independent heads, each trained to predict a fixed offset (e.g., $n+2, \dots, n+N+1$). While effective during training, this limits speculative decoding to at most $N$ draft tokens. Longer drafts require either increasing $N$ or reusing a single offset-trained head autoregressively.

Reusing a fixed-offset head introduces a training--inference mismatch: the head is trained under ground-truth hidden states but, at inference, conditions on its own generated states. This distribution shift often reduces acceptance rates as draft length increases.

Nemotron-3 Super addresses this limitation by sharing parameters across multiple MTP heads during training, yielding a unified prediction head exposed to multiple offsets. This shared-weight formulation regularizes the head across prediction horizons and improves robustness to the self-generated hidden states encountered during autoregressive drafting. As a result, the same head can be applied recursively at inference to generate longer drafts with more stable acceptance behavior. While acceptance rates naturally decrease as draft length increases, the degradation is substantially milder than with independently trained offset heads. This enables more effective speculative decoding without introducing additional parameters or requiring a separate draft model.

\begin{figure}[h!]
    \centering
    
    \begin{minipage}{0.48\textwidth}
        \centering
        \includegraphics[width=\textwidth]{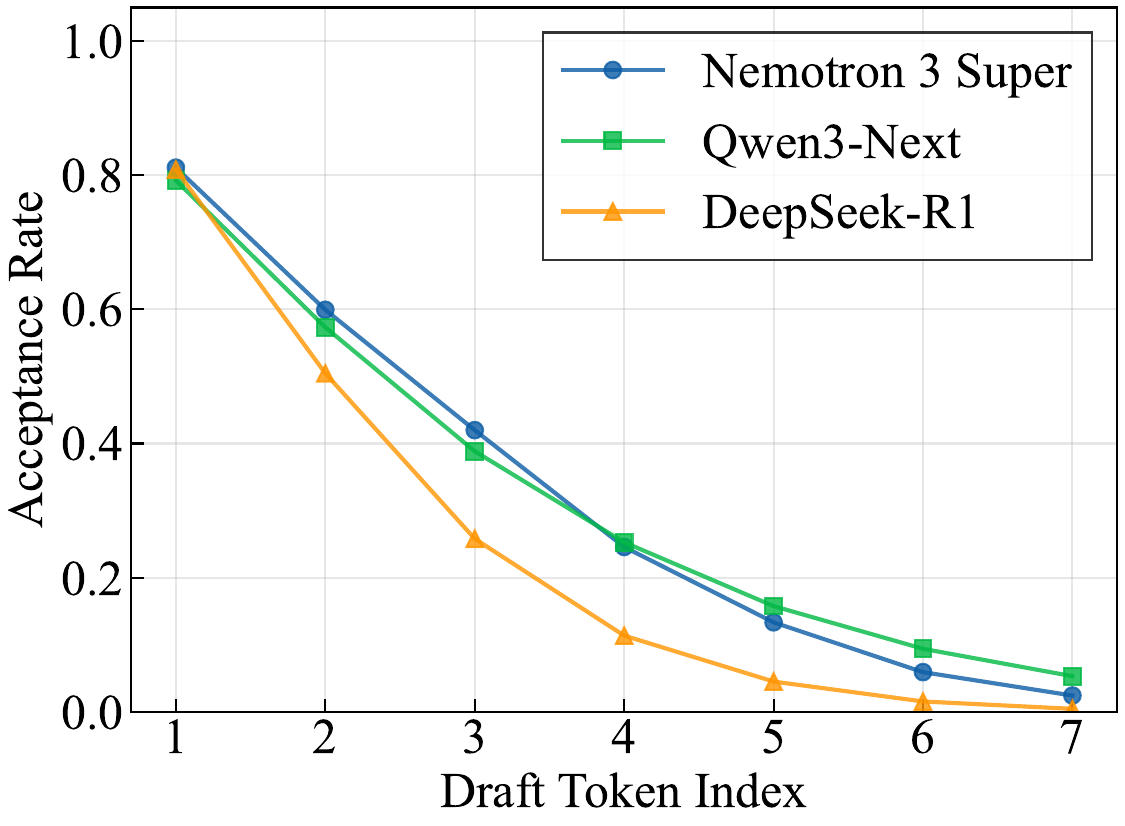}
        \caption{MTP acceptance rate by draft index on SPEED-Bench using a draft length of 7.}
        \label{fig:mtp_ar_comparison}
    \end{minipage}
    \hfill
    \begin{minipage}{0.48\textwidth}
        \centering
        \footnotesize
        \begin{tabular}{l|ccc}
        \toprule
        \textbf{Category} & \textbf{DSR1} & \textbf{\begin{tabular}[c]{@{}c@{}}Qwen3 \\ Next\end{tabular}} & \textbf{\begin{tabular}[c]{@{}c@{}}Nemotron3 \\ Super\end{tabular}} \\
        \midrule
        Coding          & 2.99 & \textbf{4.32} & 3.78 \\
        Humanities      & 2.67 & 3.07          & \textbf{3.26} \\
        Math            & 2.98 & \textbf{3.89} & 3.73 \\
        Multilingual    & 2.83 & 3.97 & \textbf{4.05} \\
        QA              & 2.63 & 3.09 & \textbf{3.16} \\
        RAG             & 2.79 & 3.53 & \textbf{3.78} \\
        Reasoning       & 2.80 & 3.47 & \textbf{3.59} \\
        Roleplay        & 2.19 & 2.17          & \textbf{2.82} \\
        STEM            & 2.79 & \textbf{3.37} & 3.30 \\
        Summarization   & 2.59 & 3.06          & \textbf{3.48} \\
        Writing         & 2.41 & 2.69          & \textbf{2.99} \\
        \midrule
        \textbf{Average} & 2.70 & 3.33 & \textbf{3.45} \\
        \bottomrule
        \end{tabular}
        \captionof{table}{MTP average acceptance lengths on SPEED-Bench using a draft length of 7.}
        \label{tab:mtp_speed_bench}
    \end{minipage} \par
    \vskip\floatsep
    \begin{minipage}{0.48\textwidth}
            \centering
        \includegraphics[width=\textwidth]{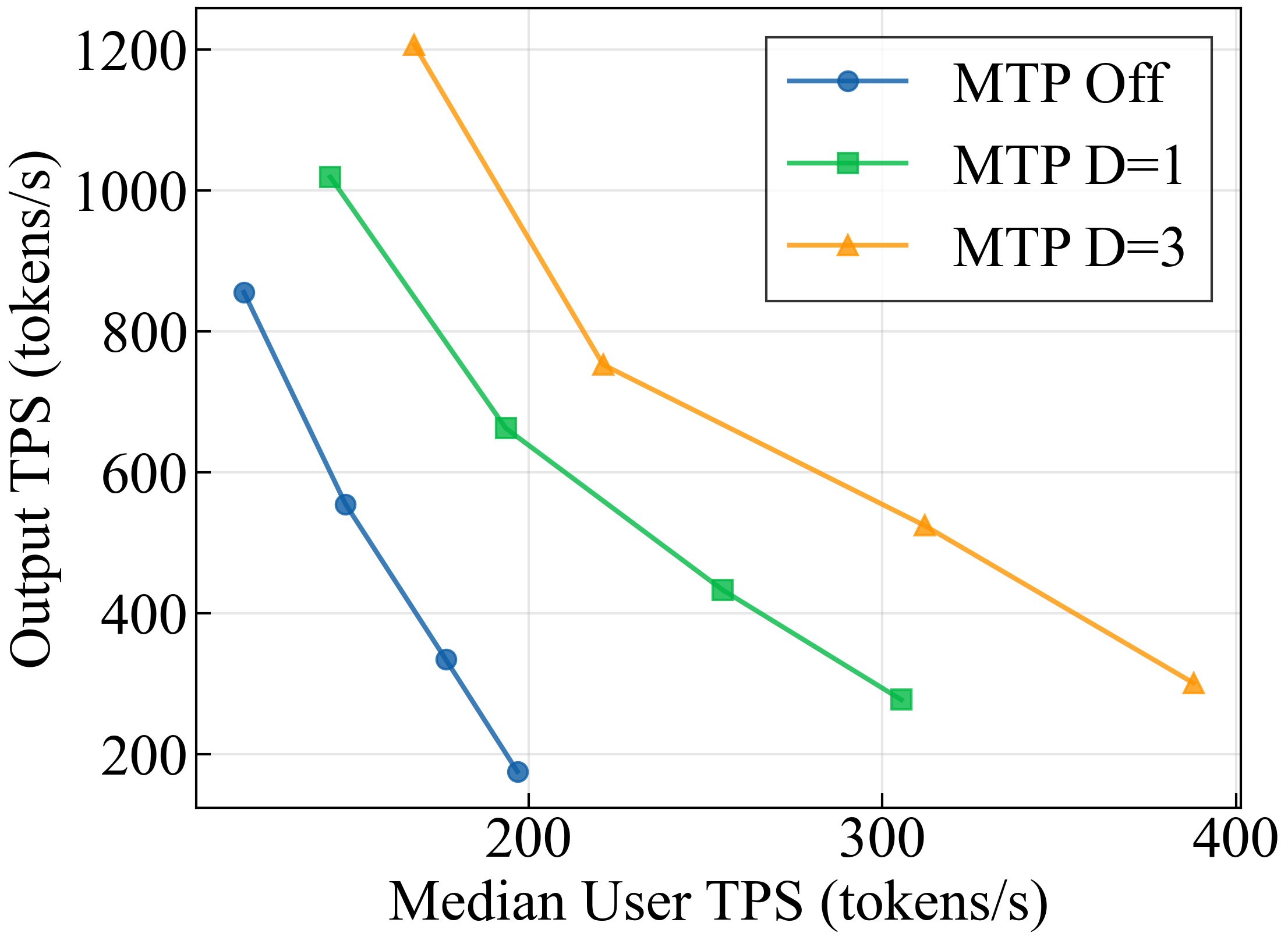}
        \caption{Total vs. user throughput for an NVFP4 checkpoint (TRT-LLM, TP=1, B300 GPU). Comparing MTP off vs. MTP with draft lengths of 1,3. Measured on SPEED-Bench's Throughput-1k split, with 1k output tokens. 
        }
        \label{fig:mtp_speed}
    \end{minipage}
\end{figure}

\paragraph{Speculative Decoding Performance.}
We evaluate MTP quality using SPEED-Bench~\citep{speedbenchnvidia}, a benchmark tailored for speculative decoding. Table~\ref{tab:mtp_speed_bench} reports the average acceptance length (tokens accepted per verification step) with a fixed draft length of 7. Nemotron-3 Super achieves the highest overall average acceptance length (3.45), outperforming DeepSeek-R1 across all domains and remaining competitive with Qwen3-Next.

Figure~\ref{fig:mtp_ar_comparison} plots acceptance rate as a function of draft token index. As expected, acceptance decreases monotonically with draft depth for all models. However, Nemotron-3 Super consistently maintains higher acceptance than DeepSeek-R1 at every draft position and closely tracks or exceeds Qwen3-Next across most indices. Notably, the gap becomes more pronounced at larger draft indices (4–7), where recursive drafting is most challenging. This behavior indicates improved stability of the shared-head autoregressive design under longer speculative rollouts.

Overall, MTP in Nemotron-3 Super improves both representation learning and decoding efficiency, enabling higher acceptance at extended draft lengths without relying on an external draft model. 
These acceptance gains translate to superior serving efficiency on Blackwell hardware. As shown in Figure~\ref{fig:mtp_speed}, increasing the draft depth ($D=1$ to $D=3$) via MTP significantly shifts the throughput--latency Pareto frontier, delivering higher aggregate output tokens per second (TPS) for any given median user latency compared to the baseline with MTP disabled. 

\subsubsection{Hybrid Interleaved MoE Architecture and Global Anchors} \label{sec:hybrid_moe_arch}

\ourmodel adopts a hybrid Mixture-of-Experts (MoE) architecture designed to maximize inference throughput—particularly for long-context reasoning—while preserving the modeling capacity of large-scale dense Transformers. The primary systems bottleneck in modern sequence models is the quadratic growth of the KV cache in self-attention layers. To address this, we predominantly utilize Mamba-2 blocks~\citep{dao2024transformersssmsgeneralizedmodels}, which operate with a constant-sized state during generation, substantially reducing memory overhead and latency.

The 88-layer stack follows a periodic interleaving pattern in which MoE layers are paired with Mamba-2 blocks. While Mamba provides efficient linear-time sequence modeling, a limited number of self-attention layers are strategically inserted as global ``anchors'' to enable full-token interaction and long-range information routing across the stack. This hybrid interleaving preserves global dependency modeling while offloading the majority of computation to the more efficient Mamba and sparse MoE components. Table~\ref{table:model_arch} and Figure~\ref{fig:layer-pattern} provide a comprehensive summary of the structural parameters and the specific interleaving pattern of the hybrid stack.

The attention layers employ Grouped-Query Attention (GQA) with 32 query heads and 2 KV heads (head dimension 128). Consistent with prior Nemotron models, we omit positional embeddings, dropout, and bias terms in linear layers, use RMSNorm for normalization, and maintain un-tied embedding and output weights. This configuration supports context lengths of up to 1M tokens.

Sparse scaling further improves efficiency. Each MoE layer activates only a subset of experts per token (top-22 routing), enabling the model to scale to 120.6B total parameters while maintaining a 12.7B active parameter budget per forward pass. 

Overall, the synergy between linear-time Mamba blocks, sparsely activated MoE capacity, and strategically placed attention anchors enables \ourmodel to deliver strong long-context performance while remaining optimized for real-world deployment on modern hardware.

%% file: sections/pretraining/nvfp4.tex
\subsection{NVFP4 Pretraining}\label{sec:nvfp4}
\begin{table*}[!hbt]\small\centering
\renewcommand{\arraystretch}{1.3}
\setlength{\tabcolsep}{10pt}
\caption{Precision by Layer Type}
\label{table:precision_by_layer}
\begin{tabular}{l|c|p{6cm}}
\toprule
\textbf{Layer Type} & \textbf{Format} & \textbf{Rationale}\\
\midrule
All Linear Layers Unless Otherwise Noted & NVFP4 &\\
\midrule
Final 15\% of Network & BF16 & Promote training stability at scale\\
\midrule
Latent Projections & BF16 & Strategically kept in BF16 as step-time impact is negligible\\ 
\midrule
MTP Layers & BF16 & Preserves multi-token prediction capabilities \\
\midrule
QKV \& Attention Projections & BF16 & Maintain fidelity of few attention layers\\
\midrule
Mamba Output Projection & MXFP8 & Mitigates high incidence of underflows observed when quantizing this layer to NVFP4 at smaller scales \\
\midrule
Embedding Layers & BF16 &\\ 
\bottomrule
\end{tabular}
\end{table*}

Nemotron 3 Super was trained with the NVFP4 pretraining recipe detailed in the Nemotron 3 white paper \citep{nvidia2025nvidianemotron3efficient}. All linear layers, unless otherwise noted in Table \ref{table:precision_by_layer}, are trained using the open-source NVFP4 GEMM kernels provided by Transformer Engine with the cuBLAS backend \citep{NVIDIA_TransformerEngine_PR2177} for fprop, dgrad, and wgrad GEMMs. This framework performs quantization of weights, activations, and gradients to NVFP4 according to the scheme first introduced in \citet{nvidia2025pretraininglargelanguagemodels}. Weights are quantized to NVFP4 using two-dimensional (2D) block scaling to maintain consistency between quantized weights in the forward and backward pass. Gradients and activations are quantized to NVFP4 using one-dimensional (1D) blocks along the GEMM reduction axis. Random Hadamard Transforms (RHTs) are performed on inputs to wgrad and stochastic rounding is applied to gradient tensors. The NVFP4 format utilizes an E2M1 element format with 16-element micro-blocks, E4M3 micro-block scaling factors, and a second-level FP32 global scale. Nemotron 3 Super showcases large-scale stable training in NVFP4 up to 25T tokens.

\begin{figure}[h]
    \centering
    \includegraphics[width=0.55\textwidth]{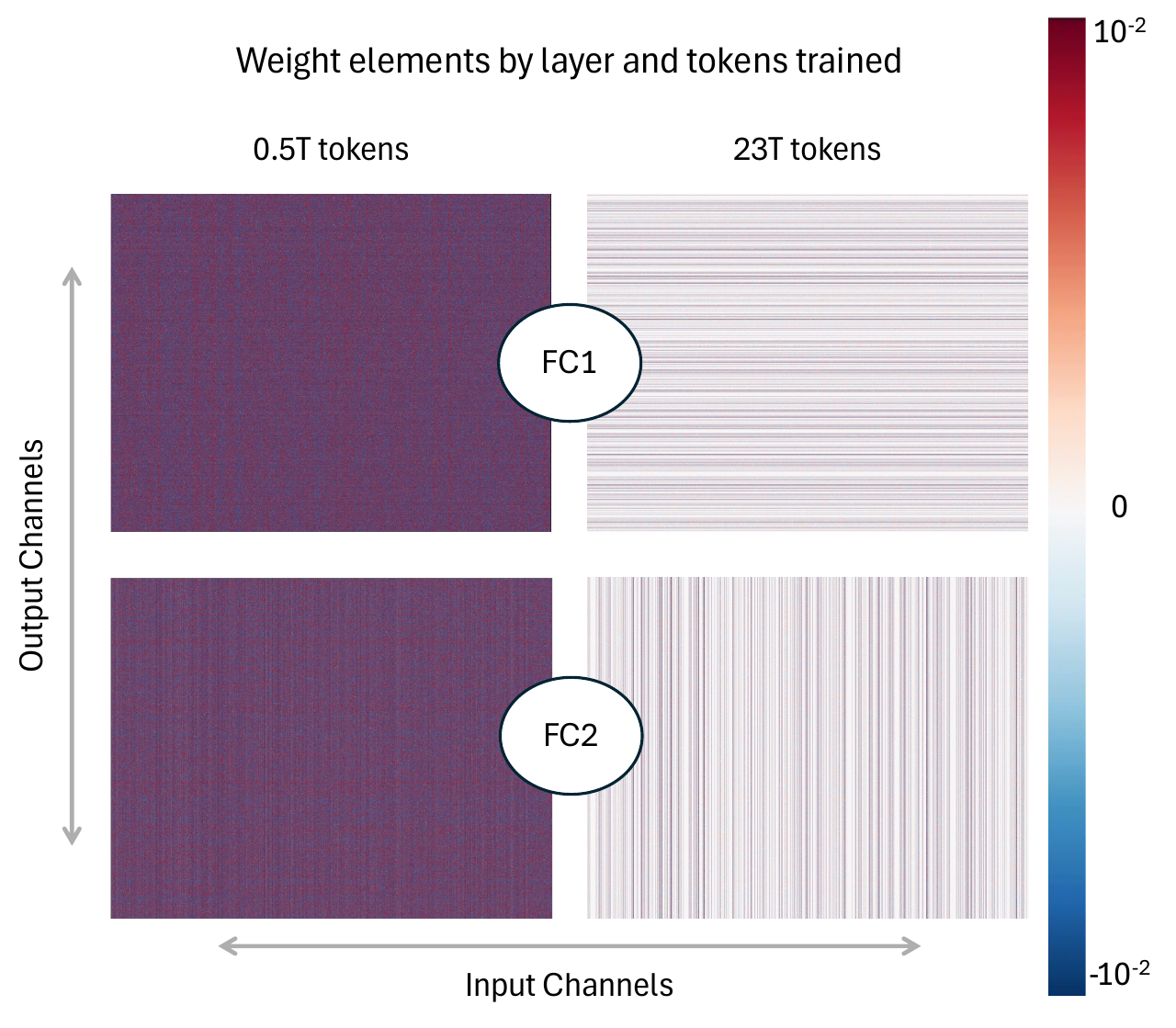}
    \caption{Channel Magnitude Patterns in Weights of Expert Layers in Nemotron 3 Super. Patterns emerge as training progresses. Top: Early layer routed expert FC1 weight matrix at 0.5T tok and 23T tok. Bottom: Early layer routed expert FC2 weight matrix at 500B tok and 23T tok. Low-norm output channels of FC1 align with low-norm input channels of FC2.}
    \label{fig:nvfp4_magnitude_patterns_super}
\end{figure}

During the training of Nemotron 3 Super, we observed a growth in the number of zero-valued weight gradient elements and investigated the root cause to validate training health. Magnitude patterns emerged within some expert layers, characterized by the norms of FC1 output channels and corresponding FC2 input channels converging toward zero (Figure \ref{fig:nvfp4_magnitude_patterns_super}). By the end of pretraining, zero-valued weight gradient elements accounted for 7\% of total parameters, appearing to correlate with the magnitude patterns. We believe that NVFP4 quantization increases the incidence of true zeros in the weight gradients that could have been more easily be represented by BF16 or MXFP8. Low-norm channels likely attenuate quicker when these layers are trained in NVFP4.

\begin{figure}[h!]
    \centering
    \includegraphics[width=1\textwidth]{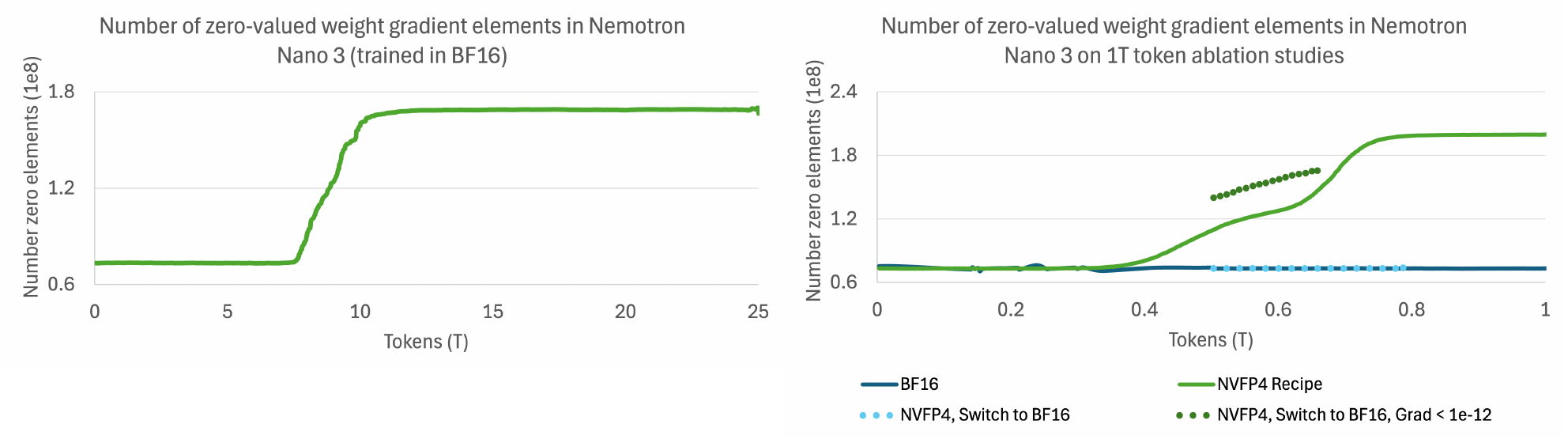}
    \caption{Number of Zero-Valued Weight Gradient Elements on Nemotron 3 Nano.
Left: Released Nemotron Nano 3 model \citep{nvidia2025nemotron3nanoopen} trained to 25T tokens in BF16. Right: Ablation study trained to 1T tokens in BF16 and with our NVFP4 recipe. The NVFP4 model at 1T tokens reaches a similar zero-valued weight gradient count as the BF16 model at 25T tokens. Switching from NVFP4 to BF16 at 0.5T tokens causes zero-valued weight gradients to return to baseline levels. The high prevalence of small-magnitude gradients (<1e-12) in BF16 suggests NVFP4 quantization underflows already-small values to zero.}
    \label{fig:nvfp4_num_zeros_in_training_nano}
\end{figure}

\begin{figure}[h!]
    \centering
    \includegraphics[width=1\textwidth]{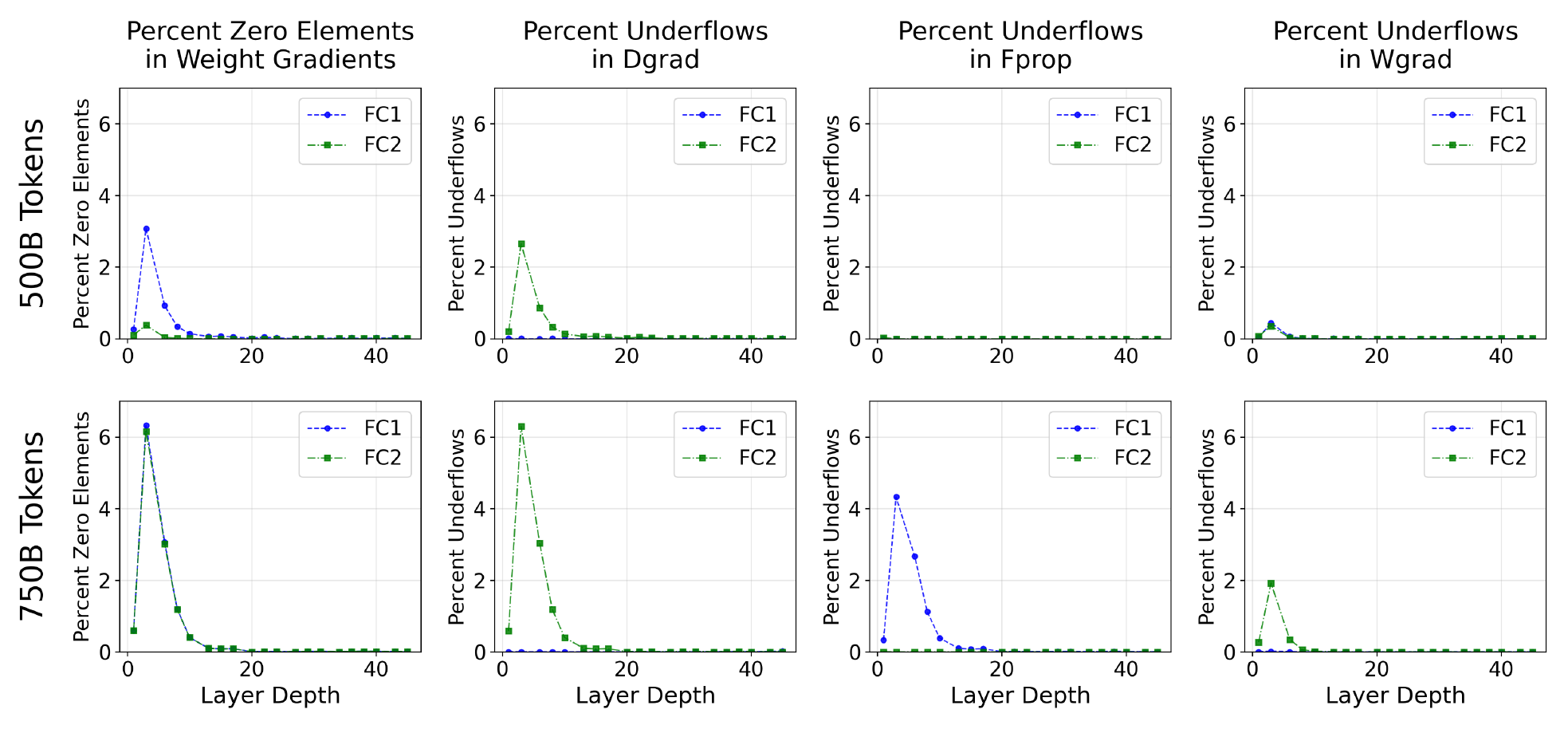}
    \caption{Origin of Zero Element Weight Gradients in Nemotron 3 Nano, shown for routed expert layers increasing in layer depth. Averaged across all routed experts in a layer index. Top: Tensors sampled at 500B tokens. Percent of zero-valued weight gradients are higher for FC1 than FC2 and attributed almost entirely to underflows in dgrad of FC2. Bottom: Tensors sampled at 750B tokens. Percent of zero-valued weight gradients are equivalently high for FC1 and FC2. Zero-valued weight gradients in FC1 are attributed to underflows in dgrad of FC2. Zero element weight gradients in FC2 are attributed mainly to underflows in fprop of FC1, with a minor contribution of underflows in wgrad of FC2.}
    \label{fig:nvfp4_wgrad_zeros}
\end{figure}

We compare identical Nemotron 3 Nano models \citep{nvidia2025nemotron3nanoopen} trained for 1T tokens in BF16 and in NVFP4 and find that NVFP4 pretraining produces roughly 3x more zero-valued weight gradients at the same token horizon. When a partially trained NVFP4 model is switched back to BF16, the number of zero-valued weight gradients returns to baseline levels. The BF16 model still contains many small-magnitude gradients (<1e-12), but NVFP4 quantization underflows these values to zero (Figure \ref{fig:nvfp4_num_zeros_in_training_nano}). We sampled weight, activation, and gradient tensors from routed expert layers and observed high rates of underflow in dgrad of FC2 at 500B tokens, primarily because two-dimensional weight quantization blocks span high and low magnitude channels. Underflows in dgrad of FC2 create zeros in wgrad of FC1 through backpropagation of the gradient. At 750B tokens, we observe high rates of underflow in fprop of FC1, creating zeros in wgrad of FC2 (Figure \ref{fig:nvfp4_wgrad_zeros}). The 1T-token NVFP4 model behaves similarly to a much longer-trained BF16 model. After 10T tokens, the released Nemotron 3 Nano model, trained in BF16, \citet{nvidia2025nemotron3nanoopen} reaches a similar number of zero-valued weight gradient elements as the NVFP4 model trained to 1T tokens (Figure \ref{fig:nvfp4_num_zeros_in_training_nano}) and inspection of weight matrices in early expert layers revealed a similar channel magnitude pattern. These conclusions on the Nemotron 3 Nano architecture give insights into the channel magnitude patterns and growth in zero-valued weight gradient elements in Nemotron 3 Super.

\begin{figure}[h!]
    \centering
    \includegraphics[width=0.65\textwidth]{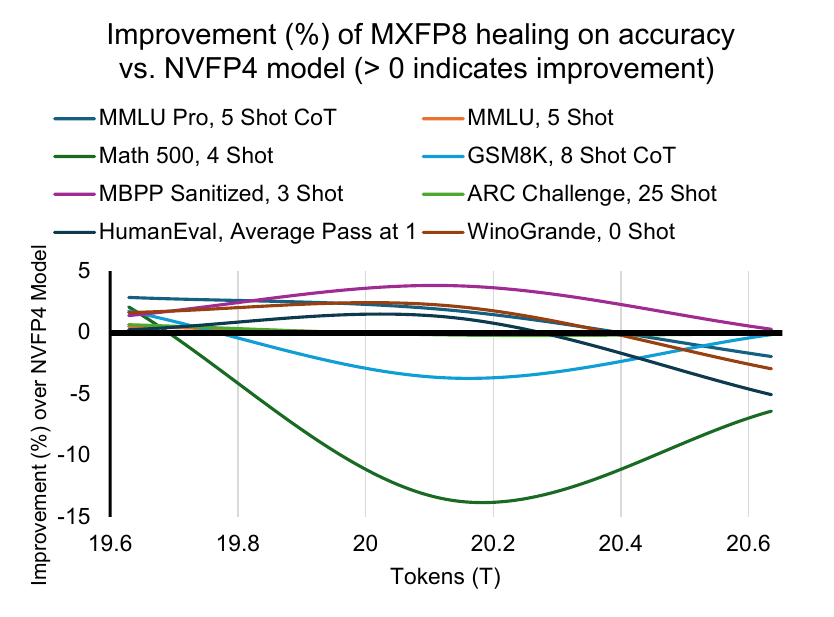}
    \caption{Improvement in downstream task evaluation accuracy after switching the network precision to MXFP8. Values greater than zero indicate improvement in accuracy over the NVFP4 model. None of the downstream task evaluation metrics showed sustained improvement after training in MXFP8.}
    \label{fig:nvfp4_healing_super}
\end{figure}

Following our previous work in NVFP4 pretraining \citep{nvidia2025pretraininglargelanguagemodels}, we evaluated whether switching all tensors to higher precision prior to learning rate annealing would benefit Nemotron 3 Super. We promoted all tensors to MXFP8 at 19T tokens (1T tokens before annealing) and continued training through 20.6T tokens. While this improved the loss trajectory, it yielded no gains in downstream task accuracy (Fig \ref{fig:nvfp4_healing_super}). The final Nemotron 3 Super model is therefore pretrained with our NVFP4 recipe for the entire token horizon.

%% file: sections/pretraining/data.tex
\subsection{Pretraining Data}
\label{sec:pretrain_data}

\subsubsection{Data}

We describe here several new datasets that we added to pretraining since Nemotron 3 Nano~\citep{nvidia2025nemotron3nanoopen}. We are releasing these datasets on HuggingFace as  \href{https://huggingface.co/datasets/nvidia/Nemotron-Pretraining-Specialized-v1.1}{\texttt{Nemotron-Pretraining-Specialized-v1.1}}.

\subsubsection{Synthetic Code Concepts}
With the aim of improving Python problem-solving capabilities, we synthetically generated a dataset consisting Python problems and solutions. Using a taxonomy consisting of thousands of programming concepts curated from our \href{https://huggingface.co/datasets/nvidia/Nemotron-Pretraining-code-v1}{\texttt{Nemotron-Pretraining-Code}} datasets and GPT-OSS-120B, we extracted high-level programming concepts from the HumanEval benchmark dataset \citep{chen2021evaluatinglargelanguagemodels}. In total, after deduplication of the extracted taxonomical representations, we collected a total of 91 concepts.

Using the extracted concepts, we performed open-ended generation using GPT-OSS 20B to generate Python programming problems that test these concepts and instructed it to generate the problem with a descriptive function name and problem description in the function docstring. To generate these problems at the pretraining scale, we combined up to four concepts per generation and generated up to five problems per set of concepts. This resulted in a total of approximately 14 million problems.

Following problem generation, we then used GPT-OSS 120B to generate five self-contained solutions for each generated problem. To avoid biasing models trained on these data to generate long-winded solutions, we instructed GPT-OSS 120B to restrict its solution to 60 lines maximum. For each problem, we generate five solutions, and we stopped generating after obtaining approximately 23 million problem-solution pairs.

As a final step in generating this dataset, we thoroughly cleaned the generated problem-solution pairs. Our cleaning consisted of the following steps: we check that GPT-OSS-120 B did not include additional imports that were not specified in the original problem generated from GPT-OSS-20 B. We discard all solutions that did not satisfy this condition.
We form the final problem-solution pair by parsing only the solution provided by GPT-OSS-120B and appending it to the problem generated from GPT-OSS-20 B. We found that GPT-OSS-120B frequently modified the original problem and this ensured we maintained the desired format prescribed in our original problem-formation prompt.
We check that the final problem-solution pair is valid Python code via generation of an abstract-syntax tree (AST). If the final function does not pass the final AST check, it is discarded.
After the above cleaning procedure, we resulted in the 15 M problems that make up the dataset.


\subsubsection{Synthetic Unconditional Algorithmic}
To create this dataset, we generated algorithmic Python problems using Qwen3-235B-A22B (the base model) and gpt-oss-120b. We used minimalistic prompts---such as ``Write a function,'' ``Write a Python function,'' or ``Write a coding problem and solution for a student to solve''---and optionally specified a difficulty level (easy, medium, or hard). To ensure diversity and quality, we prompted gpt-oss-120b to rewrite these samples to handle edge cases, add unit tests, and reformat the outputs in various ways.

In another variant, we instructed gpt-oss-120b to generate LeetCode-style questions and answers, again with a randomly selected difficulty level. We further used gpt-oss-120b to score the correctness of the solution and, if incorrect, to correct it. Overall, such nearly unconditional prompting did result in high rates of duplicates. To combat this, we found it effective to deduplicate based on short titles of around 5--8 words generated by gpt-oss-120b for each problem.

All samples were decontaminated against HumanEval~\citep{chen2021evaluatinglargelanguagemodels}, MBPP~\citep{austin2021programsynthesislargelanguage}, CRUXEval~\citep{gu2024cruxevalbenchmarkcodereasoning}, and LiveCodeBench~\citep{jain2024livecodebench} as follows: First, exact matches of the solution against those benchmarks were removed. Second, we used Qwen3-Embedding-0.6 to encode Problem and Solution and filtered any data with >0.8 similarity with any of the benchmarks.

Although this dataset is small by usual pretraining standards (0.2B tokens), we believe it helps to teach coding practices like edge case handling and reasoning about program execution, as evidenced by improvements of 1-2 points to HumanEval, MBPP, and CRUXEval-O over the Nemotron 3 Nano base checkpoint, when adding these datasets to a redo of the last 100B tokens of 25T token pretraining.

\subsubsection{Synthetic Economics}

We generated a diverse set of economics multiple-choice questions across various formats, including cloze, calculation, sentence completion, and multiple-response, covering key topics and terms in microeconomics, macroeconomics, and econometrics (e.g., ``Statistical Inference and Hypothesis Testing - Type I error'' and ``Inflation and the Price Level - Inflation rate'') from a curated list. For each topic-term pair, we used Qwen3-235B-A22B-Thinking-2507 to generate multiple questions, each accompanied by a detailed, step-by-step, and well-formatted solution. To enhance the diversity, we further prompted the model to create new and original questions using the initial outputs as reference points. Each question-solution pair underwent model-based verification for clarity, ambiguity, solvability and accuracy.

\subsubsection{Synthetic Formal Logic}

We synthesized a set of formal logic problems and solutions spanning several tasks, such as translating between natural language and predicate or propositional logic, deriving the antecedents of conditional propositions, and solving logic problems using indirect or complete truth tables.
We introduced variability into the generated scenarios, premises, and formulas by incorporating random personas,\footnote{\url{https://huggingface.co/collections/nvidia/nemotron-personas}} letters, and/or logic connective (i.e., $\land$, $\lor$, $\supset$, $\equiv$, $\sim$) into the prompt.
We generated and evaluated the problems and solutions using Qwen3-235B-A22B-Thinking-2507.

\subsubsection{Synthetic Multiple Choice}

We construct a multiple-choice question (MCQ) dataset by bootstrapping from the MMLU auxiliary training set~\citep{hendryckstest2021}, which aggregates auxiliary MCQ data from sources such as ARC~\citep{allenai:arc}, MC\_TEST~\citep{richardson-etal-2013-mctest}, OpenBookQA~\citep{OpenBookQA2018}, and RACE~\citep{lai-etal-2017-race}, etc. Starting from each seed question, we generate multiple similar questions that follow the same task format and difficulty profile, along with corresponding answer options by prompting Qwen3-235B-A22B ~\citep{yang2025qwen3technicalreport}. In the second stage, we prompt the DeepSeek-V3~\citep{deepseekai2025deepseekv3technicalreport} model to solve each generated question by selecting an answer and providing the supporting knowledge or contextual reasoning underlying its choice. To improve answer reliability, we sample multiple independent solution generations for each question using different random seeds. We then apply majority voting over the generated answers to identify the most consistent choice, retaining only those samples whose final answer agrees with the majority and discarding inconsistent or incorrect instances.

Using this pipeline, we generate approximately 3.5M MMLU-style MCQ samples (\textasciitilde1.6B tokens) augmented with explicit, relevant knowledge or reasoning traces. We evaluate the impact of this data via ablation experiments by continued training the Nemotron-Nano-V3~\citep{nvidia2025nemotron3nanoopen} 24.9T-token checkpoint with an additional 100B tokens, out of which 1B tokens are from the generated MMLU-aux-train-SDG data. The results show consistent gains on most of the benchmarks: MMLU improves from 77.22 to 77.51, "MATH Level 5" from 78.55 to 79.05, AIME-2024 improved from 53.3 to 56.7, and MBPP from from 74.8 to 75.2. Performance on other benchmarks remains largely stable with only minor variance, indicating that the synthesized MCQ data primarily strengthens mathematical and structured reasoning capabilities without introducing regressions elsewhere.

\subsubsection{Data Mixture and Ordering}

We adopt the Nemotron 3 Nano data mixture as described in~\citep{nvidia2025nemotron3nanoopen}. Our pretraining corpus spans 16 high-level categories. The largest component is web crawl data, which we partition into five quality-based groups following the Nemotron-CC taxonomy~\citep{su2024nemotroncctransformingcommoncrawl}: crawl-medium, crawl-medium-high, and crawl-high, representing progressively higher-quality crawl data, along with their synthetic counterparts, syn-crawl-medium-high and syn-crawl-high, generated from filtered web documents.
Beyond web crawl, the mixture includes math~\citep{karimi2025nemotronccmath,akter2024mindmathinformedsynthetic}, Wikipedia, code, Nemotron-CC-Code, academic text, Crawl++, multilingual data, finepdfs~\citep{kydlicek2025finepdfs} and synthetic SFT-style datasets. The SFT-style data is further divided into general-sft, stem-sft, and code-sft. As part of the SFT-style component, we incorporate reasoning-focused datasets into pretraining, motivated by prior findings demonstrating their effectiveness~\citep{akter2026frontloading}. Crawl++ consists of OpenWebText, BigScience~\citep{laurencon2023bigsciencerootscorpus16tb}, and Reddit datasets.

\begin{figure}[htbp]
    \centering
    \begin{subfigure}{0.49\textwidth}
        \centering
        \includegraphics[width=\linewidth]{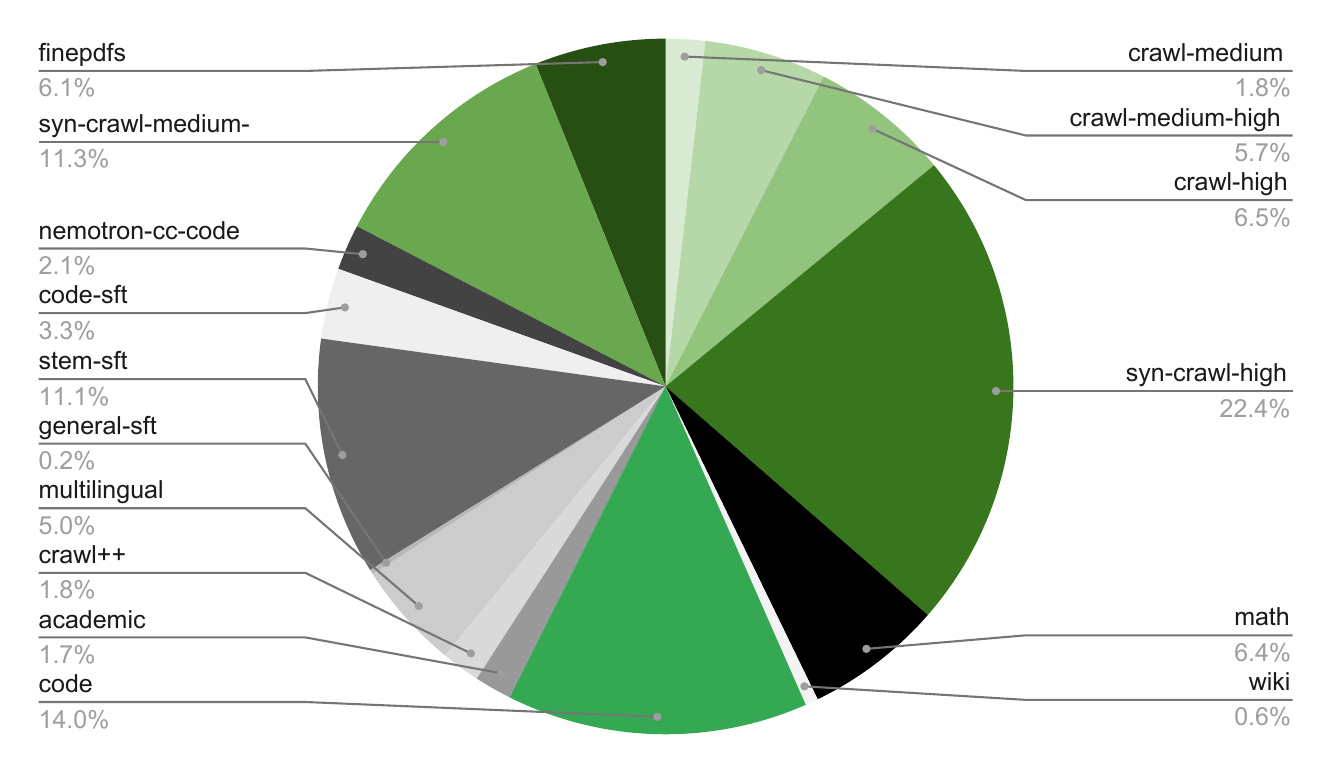}
        \caption{Data mixture of Phase 1.}
        \label{fig:phase1-blend}
    \end{subfigure}
    \begin{subfigure}{0.49\textwidth}
        \centering
        \includegraphics[width=\linewidth]{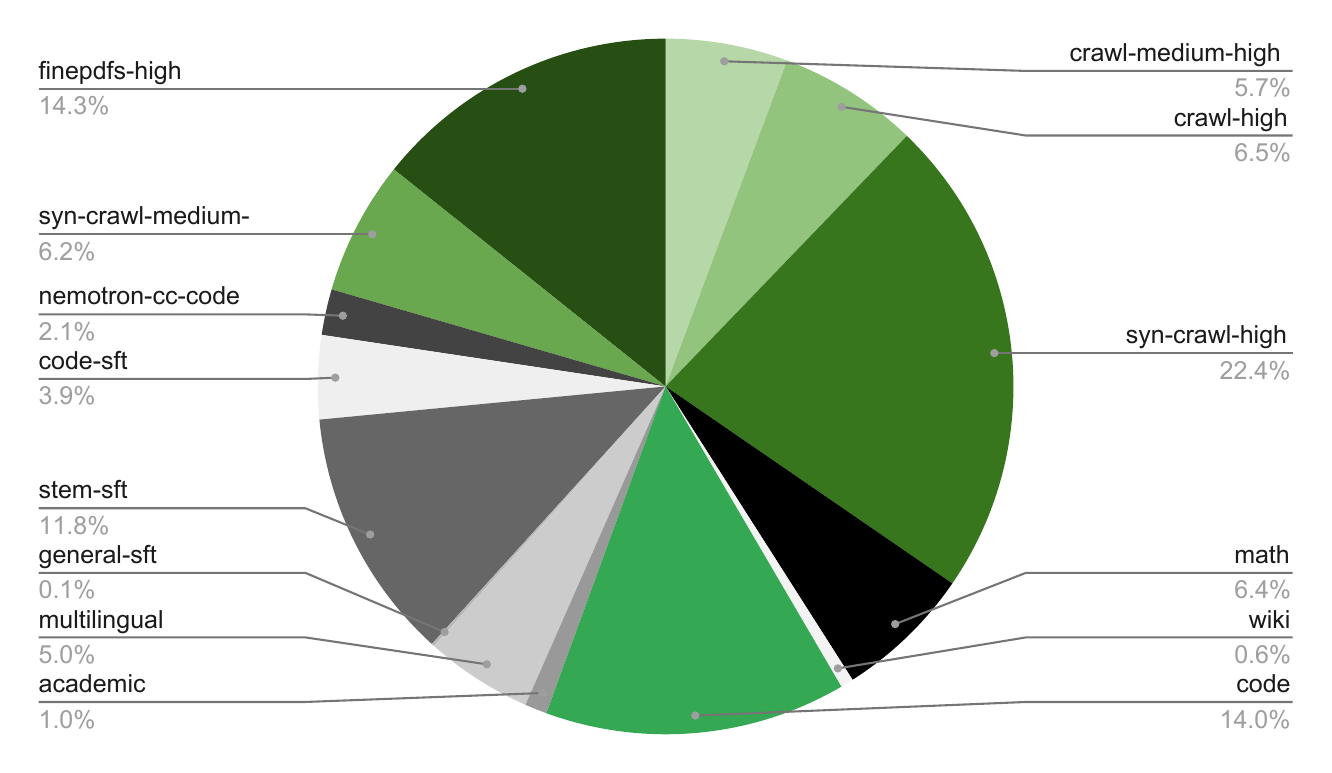}
        \caption{Data mixture of Phase 2.}
        \label{fig:phase2-blend}
    \end{subfigure}
    \caption{Data mixtures for each phase of pre-training.}
    \label{fig:pretrain-blends}
\end{figure}

Data blending is designed to balance diversity and quality: sources with comparable estimated quality are assigned similar weights, while higher-quality datasets receive proportionally greater weight in the mixture.
Further details on dataset quality estimation and mixture construction are provided in~\citep{feng2024maximizedataspotentialenhancing}.
We adopt the two-phase curriculum proposed in ~\citep{feng2024maximizedataspotentialenhancing} work. 
In Phase 1, the mixture emphasizes data diversity to promote broad coverage and generalization. 
In Phase 2, the blend shifts toward predominantly high-quality sources (e.g., Wikipedia) to refine model performance. 
The transition to Phase 2 occurs at 80\% of total training tokens. 
The specific mixtures used in each phase are illustrated in Figure~\ref{fig:pretrain-blends}.

%% file: sections/pretraining/training.tex
\subsection{Hyperparameters}\label{sec:pre-train-hyperparams}

The pretraining of \ourbasemodel was conducted using a Warmup-Stable-Decay (WSD)~\citep{hu2024minicpm} learning rate schedule over a total horizon of 25 trillion tokens. The learning rate (LR) was warmed up over the initial 200 billion tokens to a peak value of $4.5\times10^{-4}$. Following a sustained stable plateau phase, we implemented a \texttt{minus-sqrt} decay schedule for the final 5 trillion tokens, annealing the LR to a minimum of $4.5\times10^{-6}$. 

We used AdamW~\citep{loshchilov2017decoupled} optimizer with a weight decay of 0.1 and momentum coefficients $\beta_1=0.9$ and $\beta_2=0.95$. The model was trained with a sequence length of 8,192 and a batch size of 3,072 sequences, resulting in approximately 25.17 million tokens per batch.

The architecture employs a hybrid Mamba-MoE design, featuring Mixture-of-Experts (MoE) layers with 512 total experts and a top-22 routing mechanism ($k=22$). We utilized a sigmoid router score function complemented by expert biasing. To ensure equitable expert utilization across the 120.6B parameters, we adopted an auxiliary-loss-free load balancing strategy~\citep{wang2024auxiliary, deepseekai2025deepseekv3technicalreport} with an update rate of $10^{-3}$, paired with a standard load balancing loss with coefficient of $10^{-4}$~\citep{lepikhin2020gshard}. 

Furthermore, we used an MTP objective with loss scaling factor of 0.3. To maximize computational efficiency and training stability at scale, the execution utilized a hybrid precision scheme of BF16 and NVFP4.



\subsection{Tracking Merge Evaluation}\label{sec:pre-train-merge-eval}

During the stable phase of the WSD learning rate schedule described in Section~\ref{sec:pre-train-hyperparams}, the learning rate remains constant, and individual trained checkpoints exhibit noisy benchmark performance from step to step. Following recent work on weight-space merging~\citep{wortsman2022modelsoup,tian2025wsm,ling2025foundation}, we apply checkpoint merging (weighted averaging over a sliding window of recent checkpoints) to produce stronger readouts of model quality without requiring dedicated learning rate decay runs. In a conventional pretraining workflow, evaluating model quality at intermediate checkpoints requires dedicated decay runs; checkpoint merging eliminates this cost. For a schedule comparable to ours, the savings could reach ${\sim}4$T tokens of compute (e.g., ${\sim}2$ avoided runs at 1.5T and ${\sim}2$ at 0.5T), or roughly $16\%$ of the total pretraining FLOP budget.

\begin{figure}[htbp]
    \centering
    \includegraphics[width=0.9\textwidth]{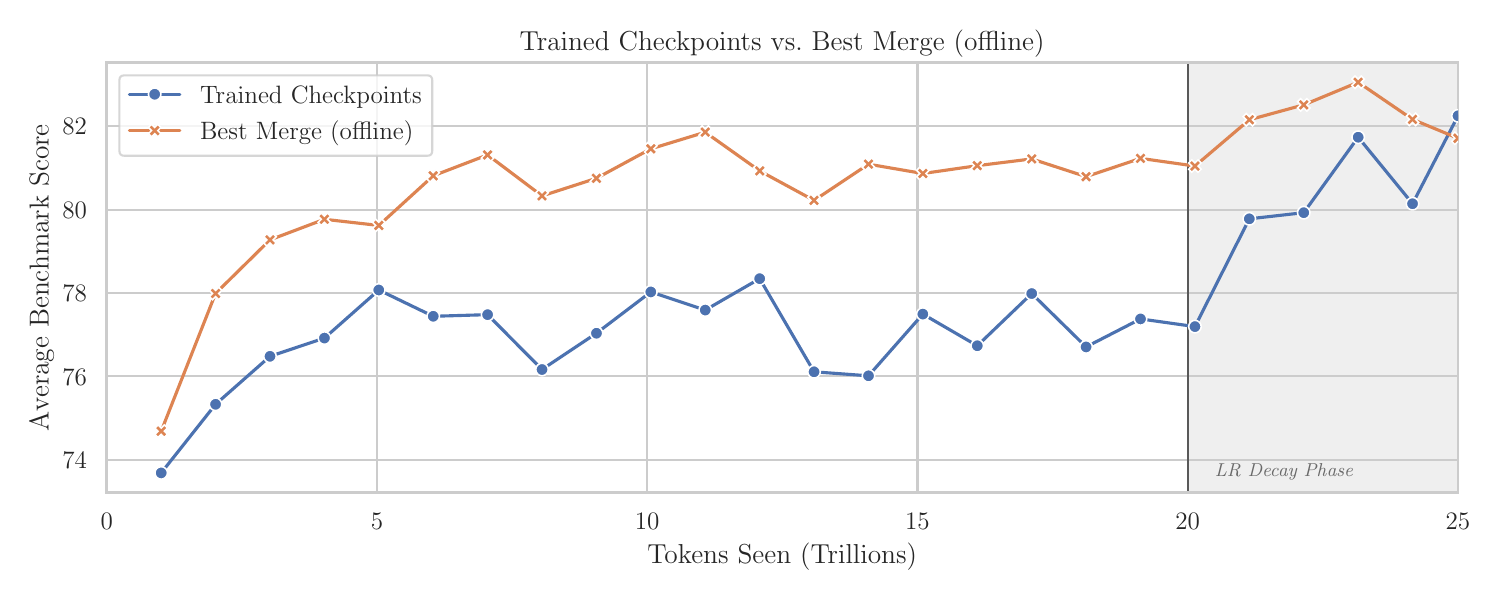}
    \caption{Average accuracy across 12 benchmarks for trained checkpoints versus the best offline checkpoint merge during pretraining. During the stable LR phase, offline merging yields a consistent 2--4 point improvement. During the LR decay phase (shaded), the gap narrows as trained checkpoints benefit from actual learning rate annealing.}
    \label{fig:avg-benchmark}
\end{figure}

Following~\citet{tian2025wsm}, we use a \texttt{minus-sqrt} decay emulation to compute merge coefficients, with checkpoints saved every $1{,}000$ iterations (${\approx}25$B tokens at our global batch size of $3{,}072 \times 8{,}192$ tokens). We evaluated sliding merge windows of 125B, 250B, and 500B tokens over the course of pretraining. On average, across a suite of 12 benchmarks (MMLU-Pro, MMLU, HumanEval, HumanEval+, MBPP, MBPP+, GSM8K, MATH-500, RACE, ARC-Challenge, HellaSwag, WinoGrande), the best merge consistently outperforms the corresponding trained checkpoint by 2--4 points on the unweighted average. Since merging is computationally cheap relative to training, we can evaluate all three windows at each checkpoint and select the best. Figure~\ref{fig:avg-benchmark} reports this best-of-three merge against the trained checkpoint over the full 25T-token training run.

During the final 5T-token LR decay phase (from 20T to 25T tokens), the gap between merged and trained checkpoints narrows substantially, and the two evaluation trajectories largely coincide by the end of training. \citet{tian2025wsm} reported that combining merging with decay offers no gain over merging alone, so this convergence is expected. The original WSM results went further, showing merge-based readouts \emph{surpassing} decay-trained checkpoints. In experiments on the Nemotron 3 Nano scale architecture (30B-A3B), we were able to reproduce such gains when emulating short (${\sim}500$B) decay windows. However, direct comparisons at 1T and 1.5T merge horizons showed no improvement over decay-trained checkpoints.

Our takeaway is that offline checkpoint merging appears most effective for shorter annealing horizons. This is consistent with~\citet{ling2025foundation}, who employ a comparably short decay schedule and report merge-based improvements, in contrast with the much longer 5T decay used here, where trained decay is able to match or surpass merge-based readouts. The final base model checkpoint selected for downstream alignment was itself a 500B merge; short-horizon merging remains practically useful even alongside a full decay schedule. That said, our experiments explored only a single merge schedule (\texttt{minus-sqrt}) and a fixed checkpoint granularity; it remains plausible that alternative coefficient schemes, finer-grained checkpoint windows, or merging strategies tailored to longer decay horizons could recover the gains observed at shorter scales. Per-benchmark breakdowns are provided in Appendix Figure~\ref{fig:per-benchmark}.

\subsection{Long-Context Extension}\label{sec:pre-train-long-context}

Similar to Nemotron 3 Nano, we added a long-context phase (LC-Phase) at the end of pretraining. In the LC-Phase, we performed continuous pretraining (CPT) to equip the base model with long-context ability. We used a constant learning rate of $4.5*10^{-6}$ and global batch size of 16. We used 64-way context parallelism, 2-way tensor parallelism, and 64-way expert parallelism to train on GB200 GPUs. We reused the long-context document QA dataset from Nemotron 2 \& 3 Nano. We allocated the document QA data to 20\% in the Phase LC data blend, with the remaining 80\% being downscaled Phase 2 data. We initially performed CPT on 1,048,576 (1m) context length. Such stage lasted for 34 billion tokens. Following that we added another stage to alternatingly train on both 1m and 4k sequences in order to mitigate the minor impact we observed on the math-related benchmarks. The second stage lasted for 17 billion tokens.

%% file: sections/pretraining/base_model_evals.tex
\subsection{Base Model Evaluations}
\label{subsec:base_model_evals}

\begin{table}[h!]
\centering
\small
\setlength{\tabcolsep}{7pt}
\renewcommand{\arraystretch}{1.3}

\begin{tabular}{lr|ccc}
\toprule
\textbf{Task} & \textbf{Metric} &
\multicolumn{1}{c}{\textbf{N-3-Super}} &
\multicolumn{1}{c}{\textbf{Ling-flash}} &
\multicolumn{1}{c}{\textbf{GLM-4.5}} \\
& &
\multicolumn{1}{c}{\textbf{120B-A12B-Base}} &
\multicolumn{1}{c}{\textbf{base-2.0}} &
\multicolumn{1}{c}{\textbf{Air-Base}} \\
\midrule

\rowcolor{black!5}
\multicolumn{5}{l}{\textbf{General Knowledge}} \\
MMLU & \textit{5-shot, acc} & \textbf{86.01} & 81.00 & 81.00 \\
MMLU-Pro & \textit{5-shot, CoT EM} & \textbf{75.65} & 62.10 & 58.20 \\
AGIEval-En & \textit{3/5-shot, CoT EM} & \textbf{77.92} & 61.70 & 62.40 \\
GPQA-Diamond & \textit{5-shot, CoT EM} & \textbf{60.00} & 36.00 & 23.20 \\
\midrule

\rowcolor{black!5}
\multicolumn{5}{l}{\textbf{MATH}} \\
GSM8K & \textit{8-shot, EM} & 90.67 & \textbf{90.75} & 82.60 \\
MATH & \textit{4-shot, EM} & \textbf{84.84} & 63.80 & 50.36 \\
MATH Level 5 & \textit{4-shot, EM} & \textbf{70.00} & 39.80 & 26.30 \\
AIME 2024 & \textit{pass@32} & \textbf{53.33} & 30.00 & 20.00 \\
\midrule

\rowcolor{black!5}
\multicolumn{5}{l}{\textbf{Code}} \\
HumanEval & \textit{0-shot, pass@1 n=32} & \textbf{79.40} & 70.10 & 76.30 \\
MBPP-Sanitized & \textit{3-shot, pass@1 n=32} & \textbf{78.38} & 77.30 & 77.50 \\
\midrule

\rowcolor{black!5}
\multicolumn{5}{l}{\textbf{Commonsense Understanding}} \\
ARC-Challenge & \textit{25-shot, acc\_norm} & \textbf{96.08} & 94.80 & 93.90 \\
HellaSwag & \textit{10-shot, acc\_norm} & \textbf{88.97} & 84.69 & 87.70 \\
OpenBookQA & \textit{0-shot, acc\_norm} & \textbf{50.20} & 47.00 & 48.60 \\
PIQA & \textit{0-shot, acc\_norm} & \textbf{85.47} & 84.00 & 84.22 \\
WinoGrande & \textit{5-shot, acc} & 78.93 & 78.37 & \textbf{83.82} \\
\midrule

\rowcolor{black!5}
\multicolumn{5}{l}{\textbf{Reading Comprehension}} \\
RACE & \textit{0-shot, acc} & \textbf{91.00} & 90.10 & 89.50 \\
\midrule

\rowcolor{black!5}
\multicolumn{5}{l}{\textbf{Multilingual}} \\
MMLU Global Lite & \textit{5-shot, avg} & \textbf{85.72} & 74.94 & 79.25 \\
MGSM & \textit{8-shot, avg} & \textbf{87.47} & 82.73 & 80.33 \\
\midrule

\rowcolor{black!5}
\multicolumn{5}{l}{\textbf{Long Context}} \\
RULER 64K & \textit{0-shot} & \textbf{92.26} & 72.12 & 80.26 \\
RULER 128K & \textit{0-shot} & \textbf{88.26} & 52.03 & 61.70 \\
RULER 256K & \textit{0-shot} & \textbf{84.56} & - & - \\
RULER 512K & \textit{0-shot} & \textbf{82.49} & - & - \\
RULER 1M & \textit{0-shot} & \textbf{71.00} & - & - \\
\bottomrule
\end{tabular}
\caption[Base model comparison]{
    Comparison of \textbf{Ling-flash-base-2.0}, \textbf{GLM-4.5-Air-Base}, and \textbf{Nemotron Super 120B-A12B Base}. Best results are marked in bold.
}
\label{tab:base-model-comparison}
\end{table}

All evaluation results were collected via Nemo Evaluator SDK\footnote{\url{https://github.com/NVIDIA-NeMo/Evaluator}} and NVIDIA's open source container of LM Evaluation Harness\footnote{\url{https://github.com/EleutherAI/lm-evaluation-harness}}, unless otherwise stated. For reproducibility purposes, more details on the evaluation settings can be found in the Nemo Evaluator SDK examples folder\footnote{\url{https://github.com/NVIDIA-NeMo/Evaluator/tree/main/packages/nemo-evaluator-launcher/examples/nemotron/nemotron-3-super}}. 
The open source container on LM Evaluation Harness packaged via NVIDIA's Nemo Evaluator SDK used for evaluations can be found here\footnote{\url{https://catalog.ngc.nvidia.com/orgs/nvidia/teams/eval-factory/containers/lm-evaluation-harness}}. This container is is built on top of LM Evaluation Harness, with the following changes applied to all models for fairness:
\begin{enumerate}
  \setlength{\itemsep}{0.4em}  
    \item For mathematical reasoning, we evaluate GSM8K and MATH~\citep{cobbe2021trainingverifierssolvemath, hendrycks2021measuringmathematicalproblemsolving} benchmarks using greedy-decoding. We also highlight the competition-level slice of the MATH benchmark as ``MATH Level 5''. Additionally, we report the $\text{pass}@32$ performance on AIME-2024. We use \texttt{Math-Verify}\footnote{\url{https://github.com/huggingface/math-verify}.} to grade all generations.
    \item For code tasks (HumanEval~\citep{chen2021evaluatinglargelanguagemodels}, MBPP~\citep{austin2021programsynthesislargelanguage}) we evaluate the EvalPlus variants along with the sanitization of generations~\citep{Liu_Is_Your_Code_2023}, in a 0-shot setup. We estimate $\text{avg}@32$,  $\text{pass}@1$ from 32 generations per prompt.
    \item General reasoning benchmarks (OpenBookQA~\citep{mihaylov2018suitarmorconductelectricity}, PIQA~\citep{bisk2019piqareasoningphysicalcommonsense}, Hellaswag~\citep{zellers2019hellaswagmachinereallyfinish}, Winogrande~\citep{sakaguchi2019winograndeadversarialwinogradschema}) are unchanged except for ARC-Challenge~\citep{Clark2018ThinkYH}, where we present all options at the same time, similar to MMLU~\citep{hendrycks2021measuringmassivemultitasklanguage}.
    \item For multilingual capability, we evaluate MGSM~\citep{shi2022languagemodelsmultilingualchainofthought} (8-shot, native CoT) and Global MMLU-Lite~\citep{singh2024globalmmluunderstandingaddressing}.
    \item For long context capability, we evaluate RULER~\citep{hsieh2024ruler} using 100 samples per task.
\end{enumerate}

Accuracy results for \ourbasemodel with comparsions to Ling-flash-Base-2.0 and GLM-4.5-Air-Base are shown in Table~\ref{tab:base-model-comparison}.


%% file: sections/alignment.tex
\section{Post-Training}
\label{sec:alignment}

\begin{figure}[h]
    \centering
    \includegraphics[width=\linewidth]{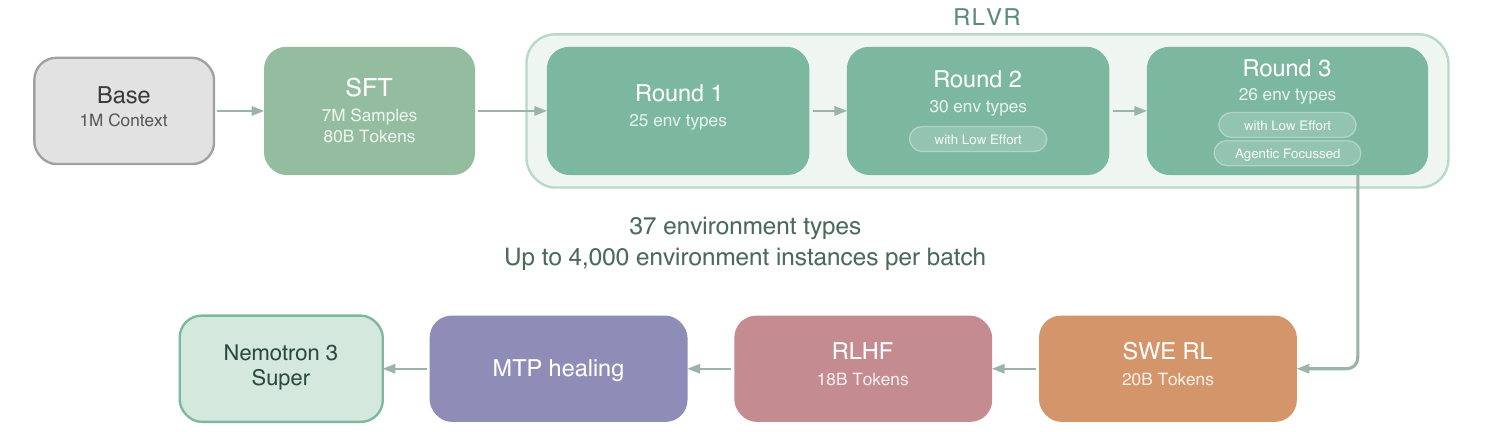}
    \vspace{-0.5em}
    \caption{Overview of the post-training pipeline for \ourmodel.}
    \label{fig:posttraining_stages}
    \vspace{-0.5em}
\end{figure}

We follow the same general recipe as Nemotron 3 Nano, with a stronger emphasis on agentic tasks. Figure~\ref{fig:posttraining_stages} provides an overview of the pipeline. We begin with a Supervised Fine-Tuning (SFT) phase (\S\ref{subsec:SFT}), followed by a three-stage Reinforcement Learning phase—RLVR, SWE-RL, and RLHF (\S\ref{subsec:RL}). We conclude with a final phase of MTP healing.

In SFT, we expand the training blend to cover a wider range of agentic harnesses and interaction scenarios. We also significantly improved our RL infrastructure, enabling reliable large-scale asynchronous training on thousands of GPUs. This infrastructure allows us to (1) train across 21 diverse environments, improving robustness across tasks, and (2) train on long-horizon SWE tasks, strengthening multi-step reasoning and problem solving in realistic agentic settings.

\input{sections/posttraining/sft}

\input{sections/posttraining/rlvr}

\input{sections/posttraining/rlhf}

\input{sections/posttraining/evals}

%% file: sections/posttraining/sft.tex
\subsection{Supervised Fine Tuning}
\label{subsec:SFT}
For \ourmodel SFT, we focused on improving dataset quality and diversity. In particular, we scaled up our agentic datasets and increased their share in the overall SFT blend. The chat template remains identical to \nanomodel. In addition, we add low effort reasoning mode, giving users further control over reasoning length. We found that a single-stage SFT led to a marked degradation on long-input-short-output scenarios. We therefore adopt a two-stage SFT procedure: Stage~1 emphasizes learning from token-level supervision and induces strong reasoning behavior, while Stage~2 switches to per-conversation normalization to prevent long outputs from dominating the loss, which restores long-input-short-output performance while retaining reasoning. We describe it below:

\subsubsection*{SFT objective and two-stage loss}

For a packed global batch $\mathcal{B}$ containing multiple conversations $c$, let $\mathcal{O}_c$ denote the set of output-token positions for conversation $c$ and $|\mathcal{O}_c|$ its output-token count. With token-level negative log-likelihood $\ell_t = -\log p_\theta(y_t \mid x, y_{<t})$, we use:

\paragraph{Stage 1: token-level (global) average.}
We minimize the average loss over \emph{all} output tokens in the packed global batch:
\begin{equation}
\mathcal{L}_{\text{tok}}
= \frac{\sum\limits_{c \in \mathcal{B}} \sum\limits_{t \in \mathcal{O}_c} \ell_t}
{\sum\limits_{c \in \mathcal{B}} |\mathcal{O}_c|}.
\end{equation}
This corresponds to summing the output-token log probabilities across all conversations and normalizing by the total number of output tokens.

\paragraph{Stage 2: sample-level average.}
We then switch to a per-conversation normalized loss and average equally across conversations:
\begin{equation}
\mathcal{L}_{\text{samp}}
= \frac{1}{|\mathcal{B}|} \sum\limits_{c \in \mathcal{B}}
\left(
\frac{1}{|\mathcal{O}_c|} \sum\limits_{t \in \mathcal{O}_c} \ell_t
\right).
\end{equation}
This stage reduces the dominance of long outputs by normalizing each conversation by its own output-token count before averaging across the batch.

For Stage 1, we run SFT with 256k sequence length packing, global batch size 64, constant lr $1e-5$ with 30k warmup samples.
 For Stage 2, we use 512k sequence length packing and include long context data with length up to 512K, global batch size 32 and constant lr $1e-5$. 

\paragraph{MTP during SFT} We continue training \ourmodel with the same shared-weight MTP head used in pretraining to preserve both the accuracy benefits of multi-step prediction and the inference-time gains from speculative decoding. Concretely, we train two MTP layers with shared parameters and optimize the combined objective using a scaled auxiliary loss computed with per-token loss and 0.3 scaling factor.

\subsubsection{Data}

We reuse the following datasets from the \nanomodel SFT datasets: Chat, Infinibyte,  and Formal Proofs. We refresh the following datasets with new teacher models (\texttt{DeepSeek v3.2, Kimi K2}): Competition Math, Competition Code,  Conversational Tool Use, Multilingual, Science. Below we describe new or heavily modified SFT datasets.

\textbf{Software Engineering.}
We curate a dataset of coding tasks derived from real-world GitHub issues to train \ourmodel for autonomous software engineering capabilities including code exploration, task tracking, issue reproduction and bug fixing. We use the issues and containerized execution environments from the SWE-Gym \citep{pan2025trainingsoftwareengineeringagents}, R2E-Gym \citep{jain2025r2egymproceduralenvironmentshybrid} and SWE-rebench \citep{badertdinov2025swerebenchautomatedpipelinetask} datasets. For R2E-Gym, we regenerate problem statements with \texttt{Qwen3-Coder-480B-A35B-Instruct}. We distill trajectories from the OpenHands agent harness using \texttt{Qwen3-Coder-480B-A35B-Instruct} as the teacher model.

\textbf{Agentic Programming.}

\begin{figure}[t!]
    \centering
    \includegraphics[width=\textwidth]{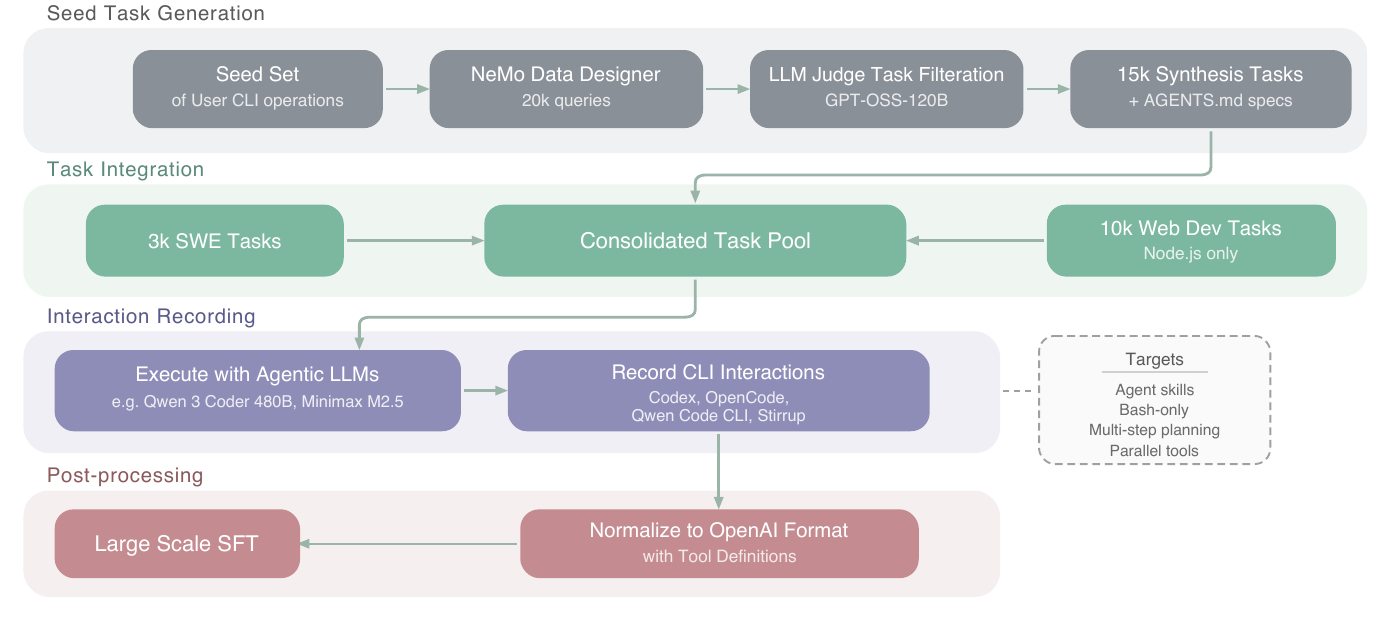}
    \vspace{-1em} 
    \caption{Agentic Command Line Interface Dataset Construction \& Training Pipeline}
    \label{figure:agentic_data_pipeline}
\end{figure}
The landscape of software development has undergone a substantial shift with the emergence of Agentic Command Line Interface (CLI) tools, moving beyond the "autocomplete" era of 2021–2023 into a regime of autonomous execution. Alongside substantial improvements in harnesses such as Claude Code, OpenCode, and OpenAI’s Codex, models are now capable of operating as active digital collaborators, capable of multi-step reasoning, long-horizon execution, and end-to-end task orchestration. 

We established a foundational seed set of tasks designed to replicate common user-initiated operations within agentic CLIs. Refer to Figure \ref{figure:agentic_data_pipeline} where we discuss the full pipeline. We utilized NeMo Data Designer \citep{nemo-data-designer} to generate approximately 20k queries derived from a taxonomy of 24 distinct actions typically performed in these environments. Subsequently, we employed \texttt{GPT-OSS 120B} \citep{openai2025gptoss120bgptoss20bmodel} in an LLM-as-a-Judge framework to filter out tasks referencing pre-existing codebases or modifications to extant files. This mitigation ensures that the models do not attempt modification operations within empty directories—a scenario that frequently results in redundant and exhausted tool invocations during failed execution cycles. The resulting dataset comprises roughly 15k tasks centered on direct solution synthesis. To further enhance the diversity of the generated outputs, we coupled each task with a supplementary markdown specification, equivalent to an AGENTS.md file. These documents impose additional constraints and architectural requirements, effectively narrowing the design space and necessitating more complex, varied solutions from the agent.

Since we eliminate all tasks that require a pre-existing codebases, we augment this task set with roughly 3000 questions from SWE tasks that are challenging and come with pre-existing repository and specific git hash commit. We apply a simple prompt prior to the existing issue statements, stating that the execution environment does not have the library installed and therefore execution of unit tests should be avoided. This is a conscientious decision made to substantially reduce the engineering effort required to support per sample container based execution of each SWE task, reduce the number of tool calls and avoid large tool outputs due to multiple rounds of unit test executions. This further relaxes the constraints imposed on the agent to solve the issue, as there is no SWE specific prompt to guide the agent in solving the task. Finally, we synthesize 10k web development tasks using a taxonomy of 100 fine-grained tasks that are commonly requested by users as the seed, on which we apply LLM-as-a-Judge to eliminate tasks that require a pre-existing repository. We impose no restrictions on these tasks, but provide only a Node.js environment and expect the agent to setup and install all dependencies and plugins on its own.

Applying these task sets, we distill from high-performance, open-source agentic LLMs such as \texttt{Qwen-3-Coder-480B} \citep{qwen2025qwen25technicalreport} and \texttt{Minimax M2.5} \citep{minimax2025m2} by recording their interactions with various CLI environments, such as Codex, OpenCode, Qwen Code CLI, and Stirrup. These interaction traces are subsequently filtered, normalized into the standard OpenAI message format with various tool definitions, and utilized for large-scale SFT to effectively embed agentic operational knowledge within the model. For each Agentic CLI, we study the individual capabilities that exist and are commonly utilized. We apply the same task sets to target different capabilities, such as Agent Skills, tool restriction (bash only execution), ask user clarifying questions, single and multi step planning, static and dynamic multi turn conversations and parallel tool calling, depending on whether the specific CLI can accommodate these capabilities. 

\textbf{Long Context.}
We extend the long-context SFT dataset from \nanomodel with a more comprehensive synthetic data pipeline. To improve long-context multi-document reasoning, we construct a synthetic SFT dataset using long sequences from our pre-training blend, which contains books, papers, financial reports, code repositories, etc. We first cluster these documents by topic/domain and concatenate related documents to reach target sequence lengths, such as 128K, 256K, or 512K tokens. 
For each long-context sample, we use an LLM to generate one or more QA pairs. The prompt requires questions to involve cross-document or cross-section navigation, ensuring information is scattered rather than localized. It strictly enforces multi-hop reasoning, requiring at least 4 to 7 distinct retrieval or reasoning steps. These steps mandate computational or logical processing, preventing simple copy-pasting, and often include explicit formatting instructions. Next, we generate 8 independent reasoning traces for each context-question pair. We apply semantic majority voting to group the answers, either by exact match or via an LLM judge. From the resulting majority group, we select the answer containing the shortest reasoning trace. 
Additionally, we generate seven synthetic reasoning tasks to improve the model's ability to process context in a sequential, left-to-right manner. Specifically, synthetic snippets (generated using \texttt{Qwen3-235B-A22B-Thinking-2507}) are concatenated together to form long input context. The thinking traces are constructed by chaining rule-based reasoning steps, each of which includes relevant excerpts from the input context and tracking metadata, such as the frequency of query-related snippets.
We also construct long-context samples by concatenating records from~\citep{nvidia/Nemotron-Personas-USA} to reach the required sequence length. Questions are designed to emphasize multi-hop reasoning and information aggregation across records. Context, question, and answer are formatted using pre-defined templates, with ground-truth answers derived by executing SQL queries over the underlying records.

\textbf{Financial Reasoning.}
To construct a large-scale training corpus for financial reasoning, we employ a template-based synthetic data generation (SDG) pipeline that scales a curated seed set into hundreds of thousands of grounded question--answer pairs. The pipeline sources 565 expert-authored seed questions from the SecQue benchmark~\citep{benyoash2025secque}, a dataset of financial analysis questions anchored to SEC 10-K and 10-Q filings. These seeds are expanded combinatorially across S\&P~500 companies\footnote{\url{https://en.wikipedia.org/wiki/List_of_S\%26P_500_companies}, accessed 2025.} and fiscal years (2019--2024), where comparative questions are restricted to company pairs within the same GICS Sub-Industry to preserve semantic coherence. GPT-OSS-120B paraphrases each template instantiation, producing up to three diverse reformulations per combination. The resulting questions are mapped to relevant SEC filing sections using the original SecQue metadata, and the corresponding documents are converted to markdown with a configurable token limit. For answer generation, we adopt the GenSelect strategy~\citep{toshniwal2025genselectgenerativeapproachbestofn}: five candidate answers are sampled per question using GPT-OSS-120B with distinct random seeds, and a larger judge model (Qwen3-235B-A22B) selects the best response based on numerical accuracy, financial methodology, and logical soundness. A smaller model (Qwen3-30B-A3B) then classifies each pair as \texttt{ANSWERABLE} or \texttt{UNANSWERABLE}, retaining only those containing a complete, substantive response. Prior to supervised fine-tuning, the SDG output undergoes percentile-based outlier removal and deduplication. The resulting dataset comprises 366,243 financial Q\&A pairs with reasoning traces.

\textbf{CUDA.}
A large-scale synthetic CUDA dataset comprising 100K samples for kernel generation, repair, and optimization was constructed using a synthetic data generation pipeline based on \texttt{DeepSeek-R1} and \texttt{GPT-OSS-120B}. Seed questions were sourced from popular open-source libraries, NVIDIA library API surfaces, and BackendBench \citep{saroufim2025backendbench}. These seeds were used to generate tuples of the form (PyTorch reference, CUDA C++ kernel) and (natural language specification, CUDA C++ kernel), each accompanied by reasoning. For each seed item, multiple candidate kernels were generated and rigorously validated for correctness within an internal CUDA evaluation environment. The validated kernels were then ranked by performance, and the highest-performing kernel was retained. In addition, we collected traces from an internal CUDA agent, producing samples of the form (PyTorch reference, faulty CUDA C++ kernel, error message, corrected CUDA C++ kernel) and (PyTorch reference, slow CUDA C++ kernel, Nsight Compute log, optimized CUDA C++ kernel). Using publicly available documentation and official code samples, and following the same formulation as the CUDA-C data, we generated additional PyTorch references and corresponding CUDA-library implementations with reasoning chains, as well as aligned natural language specifications. These libraries include Thrust, CUB, cuBLAS, cuDNN, cuSPARSE, cuRAND, and cuSOLVER.

\textbf{Safety.} We have significantly enhanced our safety framework compared to \nanomodel by combining a robust prompt library with a two-stage synthetic response generation strategy. While retaining the core prompts from Nemotron Content Safety v2 \citep{ghosh-etal-2025-aegis2}, Gretel Safety Alignment v1 \citep{gretelai_gretel-safety-alignment-en-v1}, Harmful Tasks \citep{hasan2024pruning} and Red-Team-2K \citep{luo2024jailbreakv_robustness} covering content safety and common jailbreak techniques, we added in new synthetic prompts targeting the elicitation of over-refusals, demographic biases, and copyright reproduction. We also expand coverage of jailbreak strategies to better capture emerging adversarial techniques including indirect prompt injection attacks. 

Our primary advancement over \nanomodel is an explicit response policy framework. For each prompt, a response policy is inferred from a combination of prompt metadata, its annotated safety category, and predictions from a set of lightweight auxiliary classifiers that detect attributes such as self-harm risk, demographic targeting, or the presence of embedded adversarial instructions. We cast this decision process as a multi-class classification problem, where each class corresponds to a distinct response mode aligned with safety guidelines. These response modes specify whether the model should provide supportive resources such as helpline information, issue a refusal with a brief explanation, or answer the benign portion of the request while ignoring malicious content. This ensures that the responses are safe, contextually appropriate, and consistent across a diverse set of safety-sensitive scenarios.

Following the deliberative alignment framework \citep{guan2025deliberativealignmentreasoningenables}, we adopt a two-stage generation process in which the reasoning trace and the final response are curated separately but consistently with our response policy. In the first stage, we construct a concise reasoning trace that guides the model to reflect on the safety properties of the prompt, explicitly identifying why the request may be unsafe or policy-relevant and what constraints should govern the response. In the second stage, we generate the final response based on this reasoning trace, ensuring that it adheres to the predefined response policy and behavior guidelines. This structured separation encourages deliberate reflection on safety guidelines while producing responses that are consistent, policy-compliant, and contextually appropriate to ensure that the final response follows safety policies while minimizing unnecessary references to said policies. Finally, we apply a content-moderation classifier to filter any responses flagged as unsafe, providing an additional safeguard to ensure alignment with safety objectives.

\textbf{Search.}
To improve search capabilities, we generated a synthetic search-agent SFT dataset using NeMo Data Designer \citep{nemo-data-designer}. The pipeline begins by constructing seed prompts grounded in the Wikidata knowledge graph \citep{vrandecic2014wikidata}: we query SPARQL \citep{sparql2013} for well-connected hub entities across approximately 25 verified entity classes (cities, universities, films, chemical elements, etc.), then perform random walks of 4-8 hops through the graph, filtering out degenerate paths via stop-node lists, anti-meta-relation exclusions, and minimum path-length thresholds. Each valid walk yields a start entity, a chain of factual relations, and a final answer entity.

Data Designer then processes these seeds in three stages: (1) a draft stage converts the structured knowledge-graph walk into a natural-language multi-hop question, (2) an obfuscation stage rewrites the question to hide intermediate entities and eliminate breadcrumb-style chaining -- producing search-riddle queries where the solver must decompose the problem independently -- and (3) an agent stage in which MiniMax-M2 \citep{minimax2025m2} solves the obfuscated question by issuing web searches via the Tavily (\url{https://tavily.com}) MCP search tool, producing a grounded search trajectory with supporting URLs. Each resulting SFT record is a multi-turn conversation where assistant turns interleave chain-of-thought reasoning with structured tool calls, and tool-response turns return search results as JSON, preserving a full Thought--Action--Observation loop across an average of 12 tool calls per trajectory. A final structured-output stage normalizes the agent's raw response into a validated JSON schema. 

\textbf{Terminal Use.}
The dataset for enhancing terminal capabilities follows the dual-stream \textit{Terminal-Task-Gen} methodology described in Nemotron-Terminal \citep{pi2026dataengineeringscalingllm}, comprising a total of 84,864 samples. This pipeline combines the adaptation of existing high-quality datasets with synthetic task generation grounded in a comprehensive terminal skill taxonomy. The source distribution consists of 68,924 synthetic samples, 8,125 samples from Nemotron-Cascade-Math, and 7,815 samples from Nemotron-Cascade-Code \citep{wang2025nemotroncascadescalingcascadedreinforcement}. For trajectory construction, we use \texttt{DeepSeek-V3.2} \citep{deepseekai2024deepseekv32} as the primary engine to generate step-by-step solution traces within isolated, Dockerized environments through an agentic execution-feedback loop. All samples are generated using the Terminus 2 agent framework~\citep{merrill2026terminalbenchbenchmarkingagentshard} as the underlying scaffolding, providing a unified set of terminal tools and a structured interaction protocol to maintain consistency and quality across long-horizon trajectories.

\textbf{Multilingual.}
Our multilingual data combines synthetic translations of English SFT examples with a sentence-level parallel corpus to improve machine translation. We reuse the line-by-line translation pipeline from \nanomodel, translating into six languages (German, Spanish, French, Italian, Japanese, and Chinese) using \texttt{Qwen2.5-Instruct-14b}. After translation, we apply filtering to remove samples in the wrong language and other common failure modes. We observed a recurring pattern where translation disrupts the alignment between prompt specifications and answer formats — a consistency usually maintained in English data. This mismatch led to instruction-following failures during preliminary testing. To mitigate this, we introduced a lightweight post-editing step with \texttt{Qwen3-4B-Thinking-2507} to automatically restore format compliance. We also expand the parallel corpus with additional Chinese $\leftrightarrow$ English pairs and exclude very short samples that previously degraded post-training performance.

\textbf{Structured Query Language (SQL).}
To improve \ourmodel on enterprise SQL workloads, we generate a synthetic text-to-SQL dataset with NeMo Data Designer \citep{nemo-data-designer}. The dataset contains 96.5k records spanning MySQL, PostgreSQL, and SQLite across 60 industry sectors, $\sim$700 domain topics, and 90 SQL concept buckets (from basic SELECT to recursive CTEs, window functions, and geospatial queries). Each sample pairs a natural-language prompt and a fully synthetic database schema context with a target SQL query. To improve robustness and to mimic the real-world messiness of production databases, the pipeline injects distractor tables and columns into the database context. Specifically, related but irrelevant tables and columns are added to the database context, forcing the model to learn to ignore irrelevant schema elements. Prompt diversity is controlled along three axes -- instruction style (imperative, declarative, interrogative, contextual, abbreviated), linguistic register (formal, conversational, technical, academic, direct), and politeness level -- yielding naturalistic and varied user requests. The final dataset of 96.5k records is validated and filtered down by Data Designer from a larger dataset using per-dialect syntax validators and five LLM-as-a-critic judges.

\textbf{Conversational Tool Use.} Large-scale specialized tool-use training data has been adopted by many models to boost agentic capabilities \citep{Liu2024ToolACE,deepseekai2024deepseekv32,kimiteam2025kimik2openagentic,5team2025glm45agenticreasoningcoding}. For \ourmodel, we scale conversational tool-use data via a fully synthetic, six-stage generation pipeline:
\vspace{-0.25\baselineskip}
\begin{enumerate}
\item Domain Generation: Sample synthetic domains with a model, iteratively expanding initial generations into specialized subdomains.
\item Policy and Tool Generation: Sample customer-service policies and related tools, iteratively improving them via self-refinement; use few-shot prompting and an LM-as-a-Judge for quality filtering to maintain style and formatting.
\item Scenario Generation: Generate plausible user personas, background information, and inquiries for each policy setting.
\item Trajectory Collection: For each policy–scenario pair, simulate $16$ customer service interactions among a model-based agent, user, and environment.
\item Verification: Evaluate trajectories at both outcome and process levels using an LM-as-a-Judge.
\item SFT Data Selection: Select successful trajectories for SFT and filter for difficulty by dropping scenarios that yield all-success or all-failure outcomes.
\end{enumerate}

\begin{figure}[!t]
\centering
\includegraphics[width=0.9\linewidth]{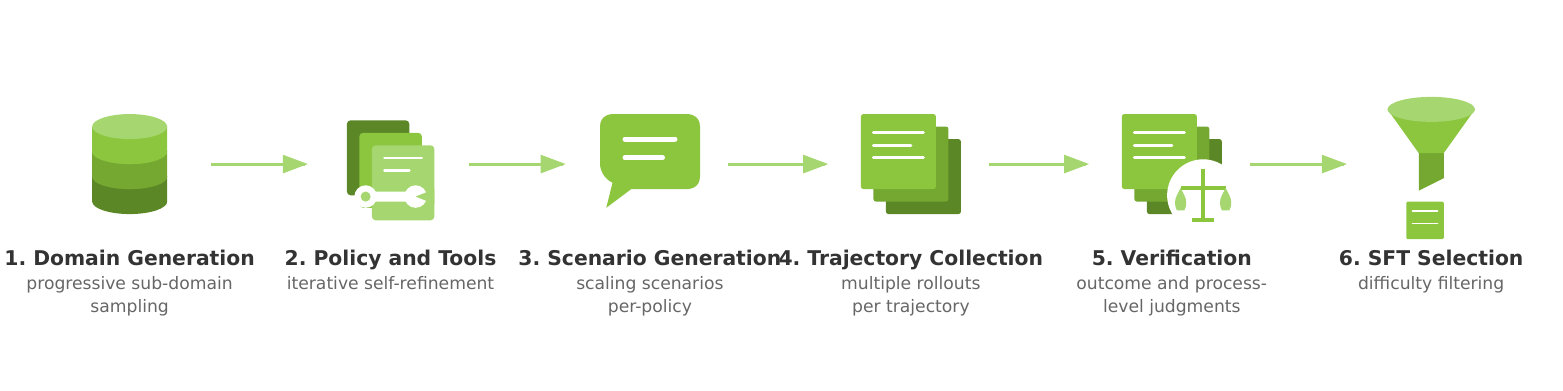}
\caption{Overview of the synthetic data generation pipeline for specialized conversational tool-use SFT data used for \ourmodel.}

\label{fig:conversational_tool_use_data_generation_pipeline}
\end{figure}

A visualization of the pipeline is shown in Figure \ref{fig:conversational_tool_use_data_generation_pipeline}. We utilize \texttt{Qwen3-235B-A22B-Thinking-2507}, \texttt{Qwen3-32B}, \texttt{Qwen3-235B-A22B-Instruct-2507} \citep{yang2025qwen3technicalreport}, \texttt{deepseek-r1-0528} \citep{deepseekai2025deepseekr1incentivizingreasoningcapability}, \texttt{DeepSeek-V3.2} \citep{deepseekai2024deepseekv32}, and \texttt{gpt-oss-120b} \citep{openai2025gptoss120bgptoss20bmodel} on various parts of the above pipeline, yielding 279{,}116 conversations across 838 domains. This represents a substantial scale-up over \nanomodel, which used 15{,}588 conversations spanning 5 domains.

\textbf{General-Purpose Tool Use.}
The broader, general purpose tool-calling synthetic data pipeline begins with the construction of diverse tool sets from ToolEyes \citep{ToolEyes2025}, API-Bank \citep{APIBank2023}, UltraTools \citep{Huang2024UltraTool}, AutoTools \citep{Shi2025AutoTools}, xLAM \citep{Zhang2024xLAM}, Glaive-Function-Calling-v2 \citep{GlaiveFunctionCallingV22025}, Toucan-1.5M \citep{Xu2025Toucan}, as well as custom written tools that serve as the foundation for downstream synthetic task generation. A tool-calling trajectory is simulated by grounding in one or multiple of these tool sets. The trajectory simulation involves an LLM playing three roles - User (User-LLM), Assistant (Assistant-LLM), and Tool Environment (Tool-LLM). The User-LLM is seeded with the selected tool set, a persona sampled from Nemotron-Personas-USA \citep{nvidia/Nemotron-Personas-USA}, and a tool-calling scenario (single-turn, multi-turn, or multi-step). The User-LLM starts by designing a task guided by the tool-calling scenario which is relevant to the selected persona and can be solved by the selected tool set. The Assistant-LLM attempts to solve this task in one or more turns by producing tool-calls and responding to tool execution results. The Tool-LLM is responsible for producing a simulated tool execution result based on the tool-call generated by the Assistant-LLM and the tool being called. The Tool-LLM is prompted with a rubric that helps identify syntactic and semantic errors in tool-calling, as well as the original user query so that the tool results can be contextualized when tool-call is successful. To ensure accuracy, we employ a turn-level and trajectory-level judge similar to the specialized tool-calling data generation. The turn level judge is also paired with a rule-based verification for ensuring correctness of tool-calls. We scale this pipeline with DeepSeek-v3.2 \citep{DeepSeekV32025} and GLM-4.7 \citep{GLM47Zai2025} to create a dataset of 1.5M diverse tool-calling trajectories.

\begin{figure}[!t]
\centering
\includegraphics[width=0.5\linewidth]{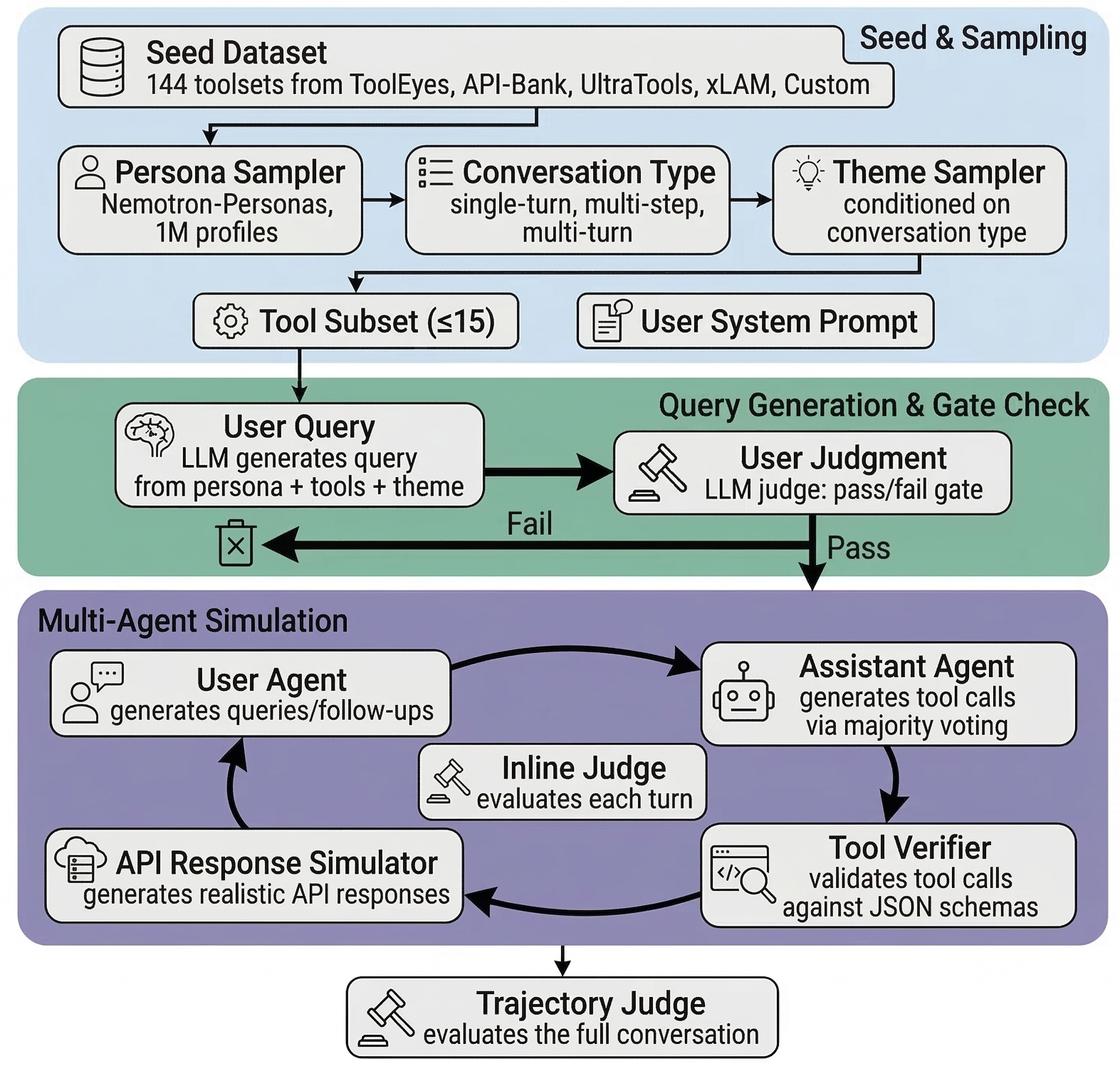}
\caption{Overview of the pipeline for general-purpose tool-calling data used in \ourmodel.}

\label{fig:broad_conversational_tool_use_data_generation_pipeline}
\end{figure}

The overall general-purpose synthetic tool-calling data pipeline is visualized in figure \ref{fig:broad_conversational_tool_use_data_generation_pipeline}.

\subsubsection{SFT Data Blend}
\begin{figure}[t]
    \centering
    \includegraphics[width=10cm]{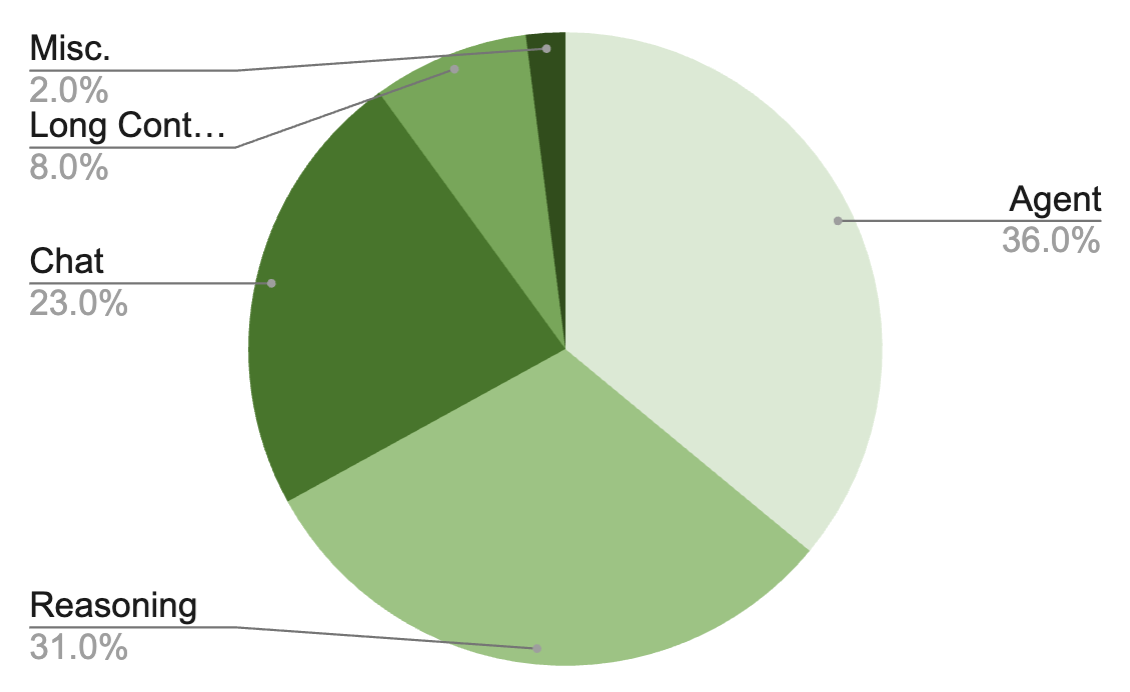}
    \caption{SFT data blend for \ourmodel. }
    \label{fig:sft_blend}
\end{figure}

Our general stage 1 SFT data blend can be found in Figure \ref{fig:sft_blend} (all datasets not listed make up less than 1\% of the blend). We train on over 7M total samples. In stage 2, we use 85\% of the stage 1 blend and augments it with 256K and 512K-token long-context data. Compared to \nanomodel, we  significantly increased the volume and diversity of agentic tasks, and allocate it a much larger proportion of our blend.

\subsubsection{Reasoning Control}

\ourmodel is trained for three reasoning modes: reasoning-off, regular and low-effort. The low-effort reasoning mode is a new addition.
The regular and low-effort reasoning modes have the option to be used in conjunction with inference-time budget control \citep{nvidia2025nemotronhfamilyaccurateefficient}.
These combinations of controls provide flexibilities that cover the entire spectrum of accuracy-efficiency trade-off to meet customers' needs in various application scenarios.

The low-effort reasoning mode is introduced during the SFT stage by adding training samples generated by GPT-OSS-120B in its low-effort mode \citep{du2025nemotron}.
These low-effort training samples cover the tasks of math reasoning, STEM question answering and instruction following, and represent 2\% of the overall SFT data by sample count.
The low-effort mode is later optimized during the RL stages to be discussed in the next section.

The SFT recipe for reasoning-off mode and for inference-time budget control is similar to \cite{nvidia2025nemotron3nanoopen} with a few differences.
We strip the reasoning traces from a random 3\% samples for reasoning-off mode.
After the main SFT stage, we add a short semi-on-policy SFT stage of 350 steps for inference-time budget control, where we collect roll-outs from the model and truncate 12\% of reasoning traces to random reasoning budgets.

%% file: sections/posttraining/rlvr.tex
\subsection{Reinforcement Learning}
\label{subsec:RL}
The RL phase of \ourmodel post-training consists of three stages followed by an MTP healing stage as shown in Figure~\ref{fig:posttraining_stages}:
\begin{itemize}
    \item[] --- \textbf{Stage 1: Multi-environment RL from Verifiable Rewards} (\S\ref{subsub:rlvr}). This is the primary training stage, where we optimize \ourmodel jointly across the full set of environments. Training in a unified mixture keeps each RL update informed by the complete environment distribution and helps prevent regressions on individual tasks over the course of training.
    \item[] --- \textbf{Stage 2: SWE-RL for end-to-end software engineering tasks} (\S\ref{subsub:swerl}). We run SWE-RL as a separate stage because SWE rollouts are substantially slower to generate and typically require longer context lengths, creating a throughput bottleneck when co-trained with shorter-horizon environments. Isolating this stage allows us to tune rollout and batching settings for long-horizon, long-context trajectories.
    \item[] --- \textbf{Stage 3: RLHF} (\S\ref{subsec:RLHF}). We finally apply RLHF as a distinct stage to improve instruction-following behavior, robustness, and overall interaction quality.
    \item[] --- \textbf{Stage 4: MTP Healing}. In this stage we train the MTP heads and keep rest of the weights frozen. We re-use the prompts from RLVR, and train the MTP head using the negative log likelihood loss similar to SFT on the generated responses. We find this stage significantly improves MTP accuracy.
\end{itemize}

We describe these stages below including the training algorithm, data and systems setup.

\subsubsection{Stage 1: Multi-environment RL from Verifiable Rewards}
\label{subsub:rlvr}

We employ a unified RLVR strategy similar to \nanomodel, but significantly scale the number of environments. We find that training on all environments simultaneously yields stable gains, whereas single-environment training leads to severe regressions on other benchmarks.

Our RLVR setting contains 21 environments covering diverse domains, including math, code, STEM, safety, chat, instruction following, long context capabilities, puzzles, and various agentic tasks. For data mixture and curriculum, we adopt an approach similar to \nanomodel: we filter out prompts where the SFT model consistently provides correct answers, then sort the remaining samples via a difficulty-based curriculum. Further details of this methodology are available in \nanomodel.

\paragraph{Low-effort Reasoning}
During the multi-environmental RL stages, we convert a subset of the prompts to be in the low-effort mode. For each low-effort prompt, the reward for a roll-out is adjusted as a function of both correctness and the number of generated tokens. The low-effort prompt mix starts with subsets of Math, STEM QA and competitive coding prompts, in total representing 2\% of all RL prompts being in low-effort mode, and is later reduced to subsets of Math and STEM QA, representing just 1\% of RL prompts. For Math and STEM QA, we randomly sample a subset. For competitive coding, we have a set of coding problems that have been withheld from SFT data and we only sample from this withheld set for low-effort coding prompts. Empirical results suggest that this data strategy provides sufficient generalization and the low-effort mode is improved across a wide set of benchmarks during the course of multi-environmental RL.

\subsubsection*{RLVR Data}
We scale up our RL data significantly compared to \nanomodel. We describe the RL datasets we used in our multi environment RL below. The majority of the RL training environments are open sourced in Nemo Gym \citep{nemo-gym}. In total, we train on 21 environments and 37 different RL datasets.

\begin{itemize}
    \item[] --- \textbf{Math.} We use the same competitive math problems as \nanomodel. For this dataset, we train both with and without a Python execution tool. We also introduce a new environment for formal proof verification.
    \item[] --- \textbf{Code.} We train on competition-style code data from \nanomodel and \citet{wang2025nemotroncascadescalingcascadedreinforcement}. We also train on single-step patch generation for software engineering tasks to prepare for end-to-end RL in the next stage.
    \item[] --- \textbf{STEM.} We include the STEM datasets from \nanomodel and add newly curated, more challenging scientific problems.
    \item[] --- \textbf{Instruction Following.} We augment the instruction-following data from \nanomodel with a new multi-challenge dataset. In this setting, the agent must follow complex user instructions, with rewards computed from a predetermined rubric.
    \item[] --- \textbf{Safety.} We add two environments: one targeting reduced over-refusals on safety-related prompts, and one improving robustness to jailbreaks. For jailbreaks, seed prompts come from our SFT data; to surface harder attacks during RL, we apply an iterative attack pipeline following PAIR \citep{chao2024jailbreakingblackboxlarge}, attacking an early SFT-only checkpoint and collecting modified prompts with high attack success rates.
    \item[] --- \textbf{Long Context.} We use the same long-context environment introduced in \nanomodel.
    \item[] --- \textbf{Agentic Tool Use.} Beyond the tool-use environments from \nanomodel, we add new environments focused on conversational tool use and terminal use.
    \item[] --- \textbf{Reasoning Gym.} We train with Reasoning Gym \citep{stojanovski2025reasoning}, enabling learning over a diverse suite of reasoning tasks.
\end{itemize}

\subsubsection{Stage 2: End-to-end RL for Software Engineering}
\label{subsub:swerl}
In the SWE-RL stage, we improve the model's ability to autonomously solve GitHub issues under diverse harnesses. Each rollout launches an Apptainer container with the target repository, runs an OpenHands agent loop to produce a code patch, and evaluates it against ground-truth tests for a binary reward. For tool diversity, we implemented OpenCode and Codex agent classes within OpenHands that match the tool formats of Claude Code and Codex CLI, reusing a single harness while varying tools and prompts at training time. This multi-harness training improves the model's generalization and performance across all target harnesses at inference time.

\subsubsection{Stage 3: Reinforcement Learning from Human Feedback}
\label{subsec:RLHF}
We follow a similar approach to \nanomodel for RLHF, training a large GenRM model to provide supervision during RL. Rather than using a vanilla GenRM, we train a principle following GenRM as in \citet{wang2025rlbff}. These principles allow us to guide \ourmodel's behavior on important domains like identity and safety related topics. Similar to \nanomodel, we use \texttt{Qwen3-235B-A22B-Thinking-2507} as the initialization for training the GenRM. 

To train the GenRM we use the Helpsteer 3 dataset \citep{Wang2025HelpSteer3Preference}, commercially friendly subsets of the lmarena-140k dataset \citep{chiang2024chatbot}, and some more recently collected human preference data. Unlike \nanomodel, we train using our GenRM throughout our multi environment RL stage and also perform a separate RLHF-only stage at the end of post training.

\subsubsection{Algorithm}
We use an asynchronous GRPO setup in which training and inference are decoupled across separate GPU devices. Inference workers continuously generate trajectories, which are stored in a rollout buffer. Once enough trajectories are collected to form a batch, the batch is sent to the training engine for a model update. We push the updated weights to the inference workers as soon as a new model version is available. Because weight updates can happen mid-rollout, a single trajectory may contain tokens produced by different model versions. We do not recompute the KV cache after updating the model weights on the inference workers. To avoid excessive policy lag, which may result in accuracy degradation, we restrict the inference workers to be at most one step behind the latest model version.

To stabilize the training and minimize off-policy effects caused by the training-inference mismatch and policy lag, we mask the importance sampling ratio computed from the training and inference logprobs \citep{Shao2024DeepSeekMath, team2025every, yao2025offpolicy}.

In multi-environment RLVR, we sample 256 prompts per step and generate 16 responses per prompt. We train with a batch size of 4096, which corresponds to a single gradient update per rollout. We begin training with a maximum generation length of 49K tokens and later increase it to 64K.

\textbf{Agentic RL - PivotRL:}
Post-training for long-horizon agentic capabilities has a tension between efficiency and accuracy. By long-horizon, we mean tasks that require many turns of interaction with an environment, such as conversational tool use, code editing, terminal interaction, and web search. SFT is cheap and simple for this task, but it often degrades performance outside of the target domain (OOD). End-to-end RL avoids that outcome to a large part, but it is costly because every update requires online interactive rollouts in complex environments. To address this, during the post-training of Super, we adopt PivotRL~\citep{yi2026pivotrl}. 

PivotRL is an assistant-turn-level RL method that addresses this tradeoff by reusing offline SFT expert trajectories during RL. It focuses training on informative turns  (called "pivots") within those SFT traces, where the policy has uncertainty over the next action, and it uses a domain-appropriate reward to match the policy's action to the expert action, so the model gets credit for similar actions rather than the exact expert action. We notice that this method greatly improves the efficiency of our agentic RL, without facing the OOD degradation issues of SFT.

We apply PivotRL for all agentic domains: including for Agentic Programming, Search, Terminal Use, and Conversational Tool Use. We will have a manuscript with more details soon.

\subsubsection{Infrastructure}
RL at the frontier of model post-training is currently defined by scaling up to an increasing diversity of tasks or environments designed for the model to learn increasingly general capabilities. Scaling RL to many environments requires a high-performance, extensible, and standardized interface for coordinating between rollouts and training. To address the scaling performance and extensibility challenges using one standard framework, we adopt NeMo Gym \citep{nemo-gym} and NeMo RL \citep{nemo-rl} for enabling large-scale RL on many different environments/verifiers.

NeMo Gym is based on the abstraction of servers. There are three core varieties of servers in Gym: (1) agents, (2) models, and (3) resources. An agent server implements the rollout kernel of a RL environment. A model server wraps an inference engine such as vLLM \citep{kwon2023efficientmemorymanagementlarge} to provide a prompt-response API, and also carefully preserves token and inference log-prob data and metadata required for RL. A resource server provides a verification API for computing rewards from a given rollout.

Our Nemotron Super 3 RLVR experiments were all based on an integrated infrastructure of NeMo RL and NeMo Gym: NeMo RL acts as the RL training loop controller, using Megatron-Core \citep{shoeybi2020megatronlmtrainingmultibillionparameter} for model training at scale, and routing all rollouts through NeMo Gym and vLLM.

NeMo RL and NeMo Gym use ray for orchestration and resource management and deploy the ray cluster on SLURM. Megatron training workers, vLLM generation workers, Gym environments and judge models are all scheduled onto a single ray cluster.

\textbf{Async RL Infrastructure: } 
All RL stages used asynchronous RL, where generation can happen independently of training which improves training efficiency by trading off how on-policy the rollouts are. The trainings used one-step off policy where each training step was earmarked for a future training step so there were no wasted rollouts. Training and generation were not collocated which simplifies deployment and avoids having to orchestrate complex memory management between the async training and generation workers.

All asynchronous RL runs also used in-flight weight updates \citep{piche2025pipelinerl} where training can update generation worker weights without waiting for the remaining ongoing rollouts to finish. The result is a single rollout can have tokens and log probabilities from differently aged policies. Enabling in-flight weight updates is critical for speeding up asynchronous RL training. We did not recompute the KV cache after in-flight weight updates.

\textbf{Resiliency: }
As we scaled up to 1k GPUs, we encountered several issues that caused intermittent failures that were not observed in smaller job shapes. These issues fell into two categories (1) hardware related and (2) software related.

We observed several hardware issues that required full restart of the job, so several optimizations were also made to improve startup time by (1) parallelizing all initialization (2) prefetching all virtual environments and binaries (3) utilize caching in upstream repos like vLLM and flashinfer.

Parallel initialization exacerbated latent race conditions with port bindings in the post-training software stack, which became a significant failure point. Several components of the post-training software required ports: (1) Ray control plane (2) vLLM workers and OpenAI servers (3) TCP rendezvous (4) NeMo Gym servers. Due to the high number of processes needing ports on a node, we hit port conflict frequently at 1K GPU scale that were all time-of-check to time-of-use (TOCTOU) race conditions. The pattern we observed was that a component would check if a port was available without claiming it exclusively, and by the time it (or another process it notified) attempted to bind, another process had already claimed the port.

\textbf{SWE-RL Infrastructure: }
Training a model on software engineering tasks requires a gym environment that can execute hundreds of concurrent agent-codebase interactions, each within an isolated sandbox, and return a reward signal derived from real test execution. In the SWE-RL environment in Nemo-Gym, each rollout launches an Apptainer container with the target repository, runs the OpenHands agent loop to produce a code patch, and runs the ground-truth tests to compute a binary reward. Rollouts are distributed across nodes using Ray with a \texttt{SPREAD} scheduling strategy. Below we describe the key components of this environment.

\begin{itemize}
    \item \textbf{Container Execution with Apptainer.} The absence of root access on our cluster restricts the use of Docker for container isolation. Instead, we use Apptainer (formerly Singularity) to run each SWE task instance in a pre-built container image (\texttt{.sif} files), providing filesystem isolation via a writable tmpfs overlay while sharing the host kernel.

    \item \textbf{OpenHands Agent Loop.} The agent loop is executed by a modified version of OpenHands, managing the full lifecycle of each interaction: initializing the runtime, presenting the problem statement, running the agent's step loop upto a configurable turn limit, extracting the git patch, and cleaning up. The agent interacts with the repository workspace through bash commands and file operations via a tmux-based session.

    \item \textbf{Memory Management.} Since Apptainer containers share host memory and processes (unlike Docker's cgroup isolation), runaway agent processes can cause OOM conditions affecting the entire node. We implemented a memory watchdog daemon that monitors the aggregate RSS of the tmux process tree and proactively kills processes inside panes when a configurable limit is exceeded, while keeping the tmux server alive for graceful recovery.

    \item \textbf{Command Blocklist.} The shared-kernel nature of Apptainer means an agent issuing \texttt{killall} or \texttt{pkill} could terminate training processes or vLLM servers on the same node. A regex-based command block list intercepts and blocks dangerous commands before execution, returning informative error messages with safer alternatives.

    \item \textbf{Harness Diversity.} To increase tool diversity during training without building separate harness integrations, we implemented OpenCode and Codex agent classes within OpenHands that match the tool input/output formats of their respective external harnesses (Claude Code and Codex CLI). Both agents plug into OpenHands' existing runtime and conversation memory, inheriting container management and observation handling. 

    \item \textbf{Serialization.} We replaced Python's standard \texttt{json} with \texttt{orjson} for serialization of HTTP payloads between the gym and the model server, as each trajectory turn carries prompt token IDs, generated token IDs, and log probabilities, resulting in large payloads that benefit from orjson's Rust-based implementation.
\end{itemize}

%% file: sections/posttraining/rlhf.tex


%% file: sections/posttraining/evals.tex
\subsection{Post-trained Model Evaluations}
\label{subsec:final_model_evals}


\begin{table*}[!htp]
\centering
\small
\setlength{\tabcolsep}{7pt}
\renewcommand{\arraystretch}{1.15}
\begin{tabular}{l|c c c}
\toprule
\textbf{Benchmark} & \textbf{N-3-Super} & \textbf{Qwen3.5-122B-A10B} & \textbf{GPT-OSS-120B} \\
\midrule

\rowcolor{black!5}
\multicolumn{4}{l}{\textbf{General Knowledge}} \\
MMLU-Pro & 83.73 & 86.70 & 81.00 \\

\midrule
\rowcolor{black!5}
\multicolumn{4}{l}{\textbf{Reasoning}} \\
AIME25 (no tools) & 90.21 & 90.36 & 92.50  \\
HMMT Feb25 (no tools) & 93.67 & 91.40 & 90.00  \\
HMMT Feb25 (with tools) & 94.73 & 89.55 & -  \\
GPQA (no tools) & 79.23 & 86.60 & 80.10 \\
GPQA (with tools) & 82.70 & - & 80.09 \\
LiveCodeBench (v5 2024-07$\leftrightarrow$2024-12) & 81.19 & 78.93 & 88.00 \\
SciCode (subtask) & 42.05 & 42.00 & 39.00  \\
HLE (no tools) & 18.26 & 25.30 & 14.90  \\
HLE (with tools) & 22.82 & - & 19.0 \\

\midrule
\rowcolor{black!5}
\multicolumn{4}{l}{\textbf{Agentic}} \\
Terminal Bench (hard subset) & 25.78 & 26.80 & 24.00 \\
Terminal Bench Core 2.0 & 31.00 & 37.50 & 18.70 \\
SWE-Bench (OpenHands) & 60.47 & 66.40 & 41.9 \\
SWE-Bench (OpenCode) & 59.20 & 67.40 & - \\
SWE-Bench (Codex) & 53.73 & 61.20 & - \\
SWE-Bench Multilingual (OpenHands) & 45.78 & - & 30.80 \\
\textbf{TauBench V2} &  &  &  \\
\quad Airline & 56.25 & 66.0 & 49.2  \\
\quad Retail & 62.83 & 62.6 & 67.80 \\
\quad Telecom & 64.36 & 95.00 & 66.00 \\
\quad Average & 61.15 & 74.53 & 61.0 \\
BrowseComp with Search& 31.28 & - & 33.89 \\
BIRD Bench & 41.80 & - & 38.25 \\

\midrule
\rowcolor{black!5}
\multicolumn{4}{l}{\textbf{Chat \& Instruction Following}} \\
IFBench (prompt) & 72.56 & 73.77 & 68.32 \\
Scale AI Multi-Challenge & 55.23 & 61.50 & 58.29 \\
Arena-Hard-V2 & 73.88 & 75.15 & 90.26 \\

\midrule
\rowcolor{black!5}
\multicolumn{4}{l}{\textbf{Long Context}} \\
AA-LCR & 58.31 & 66.90 & 51.00 \\
RULER 256k & 96.83 & 96.74 & 52.30 \\
RULER 512k & 95.22 & 95.95 & 46.70 \\
RULER 1M & 91.64 & 91.33 & 22.30 \\

\midrule
\rowcolor{black!5}
\multicolumn{4}{l}{\textbf{Multilingual}} \\
MMLU-ProX (avg over langs) & 79.36 & 85.06 & 76.59 \\
WMT24++ (en$\rightarrow$xx) & 86.67 & 87.84 & 88.89 \\

\bottomrule
\end{tabular}

\caption{
Evaluation suite for \ourmodel. We compare against \textsc{Qwen-3.5-122B-A10B} and \textsc{GPT-OSS-120B}.
}
\label{tab:super_comparison}
\end{table*}

We evaluate \ourmodel on the same broad benchmark suite and evaluation stack as \nanomodel\citep{nvidia2025nemotron3nanoopen}, covering general knowledge, reasoning, agentic, instruction following, long-context, and multilingual capability. All evaluation results were collected via Nemo Evaluator SDK\footnote{\url{https://github.com/NVIDIA-NeMo/Evaluator}} and for most benchmarks, the Nemo Skills Harness\footnote{\url{https://github.com/NVIDIA-NeMo/Skills}}. For reproducibility purposes, the open source container on Nemo Skills packaged via NVIDIA's Nemo Evaluator SDK used for evaluations can be found here\footnote{\url{https://catalog.ngc.nvidia.com/orgs/nvidia/teams/eval-factory/containers/nemo_skills}}. In addition to Nemo Skills, the evaluations also used dedicated open-source packaged containers for Tau-2 Bench (default prompt), Terminal Bench Hard (48 tasks), ScaleAI Multi Challenge Multi-turn Instruction Following, Ruler. More details on the evaluation settings can be found in the Nemo Evaluator SDK configs folder\footnote{\url{https://github.com/NVIDIA-NeMo/Evaluator/tree/main/packages/nemo-evaluator-launcher/examples/nemotron/nemotron-3-super}}. The following benchmarks are not onboarded yet in our open source tools and for these we used either their official open source implementation or otherwise an internal scaffolding that we plan to open source in the future: SWE Bench Verified (OpenHands), SWE Bench Multilingual (OpenHands), BrowseComp with Search (internal implementation, with Serp API), Terminal Bench Core 2.0 (Harbor).

\paragraph{Reasoning Capabilities.}
We report results on \textsc{AIME 25}, \textsc{HMMT Feb 25}, \textsc{GPQA} \citep{rein2023gpqa}, \textsc{LiveCodeBench v5} \citep{jain2024livecodebench}, \textsc{SciCode} \citep{tian2024scicoderesearchcodingbenchmark}, and \textsc{HLE} \citep{phan2025humanitysexam}. Across all benchmarks \ourmodel is competitive with \texttt{GPT-OSS-120B}, while lagging behind \texttt{Qwen-3.5-122B} slightly.

\paragraph{Agentic capabilities.}
 We report results on \textsc{TerminalBench} (the hard subset and the v2 set),
\textsc{SWE-Bench} (OpenHands, OpenCode, Codex, and the Multilingual set) \citep{jimenez2023swe}, \textsc{TauBench V2} (Airline, Retail, Telecom; and their average) \citep{barres2025tau},
and \textsc{BrowseComp} \citep{wei2025browsecomp}. For Browsecomp, our harness takes strong inspiration from the browser tool released with GPT OSS. \citep{openai2025gptoss120bgptoss20bmodel}, and we do not evaluate with any context management strategies. For SQL we evaluate on the BIRD benchmark \cite{li2023bird} dev set (1,534 samples, SQLite, execution accuracy). Across all agentic benchmarks, \ourmodel outperforms or is competitve with GPT-OSS 120B and is competitive to Qwen 3.5 122B on some harnesses.


\paragraph{Chat and Instruction Following Capabilities.}
We report results on \textsc{IFBench}, \textsc{Multi-Challenge}, \textsc{Arena-Hard V2} \citep{li2024wildchat}. Across all benchmarks \ourmodel is competitive with the baseline models.

\paragraph{Long Context Capabilities.}
We report results on \textsc{Ruler} \citep{hsieh2024ruler} using 100 samples per task and \textsc{AALCR}. 

\paragraph{Multilingual Capabilities.}
We measure multilingual capability on 
\textsc{MMLU-ProX} \citep{xuan2025mmlu} and \textsc{WMT24++} en$\rightarrow$xx \citep{deutsch2025wmt24++}. \ourmodel matches or outperforms the baseline models on both bencharks.

For comparison with \textsc{GPT-OSS-120B} and \textsc{Qwen-3.5-122B-A10B}, we use officially reported numbers
whenever available; when not available, we follow the \nanomodel report \citep{nvidia2025nemotron3nanoopen} procedure of either sourcing values
from reputable public aggregators (when consistent with the official protocol) or computing scores
ourselves using the official evaluation settings.

%% file: sections/quantization.tex
\section{Quantization For Inference}
\label{sec:quantization}

We apply post-training quantization (PTQ) using Model-Optimizer\footnote{\url{https://github.com/NVIDIA/Model-Optimizer/}} to quantize weights and activations to generate two efficient deployment checkpoints: FP8 (W8A8) for Hopper and NVFP4 (W4A4) for Blackwell. 

\subsection{Nemotron 3 Super FP8 Checkpoint}
For FP8 PTQ calibration, we used a small subset containing 256 samples with 65536 context length from the post-training SFT dataset. For FP8 quantization, we quantized MoE GEMMs, both routed and shared, and Mamba Linear layers. We also kept the KV Cache in FP8, whereas Mamba state cache has been quantized to FP16 for speedup. The precision assignments for the different operators in this checkpoint are summarized in Table~\ref{tab:precision}.

\begin{table}[h]
\centering
\caption{Precision settings for the FP8 checkpoint compared with the BF16 baseline.}
\label{tab:precision}
\begin{tabular}{l c c}
\toprule
\textbf{Configuration} & \textbf{FP8 Checkpoint} & \textbf{BF16 Baseline} \\
\midrule
Embedding & BF16 & BF16 \\
\midrule
Attention GEMM (QKV and Out Projection) & BF16 & BF16 \\
KV Cache + Attention BMM1 & FP8 & FP8 \\
Attention BMM2 & BF16 & BF16 \\
\midrule
MoE GEMM (Sparse Experts and Shared Experts) & FP8 & BF16 \\
MoE Latent Projection GEMM & BF16 & BF16 \\
Router & FP32 & FP32 \\
\midrule
Mamba GEMM & FP8 & BF16 \\
Mamba SSM Cache & FP16 & FP32 \\
Mamba 1D Conv & BF16 & BF16 \\
\midrule
Output Layers & BF16 & BF16 \\
\bottomrule
\end{tabular}

\vspace{2pt}
\begin{minipage}{\linewidth}
\footnotesize
\end{minipage}
\end{table}


\subsection{Nemotron 3 Super FP4 Checkpoint}

FP4 is a more aggressive quantization format than FP8 and is particularly attractive for prefill-heavy inference workloads, such as coding-agent deployments, where linear and MoE GEMMs are major performance bottlenecks. NVFP4 is natively accelerated on Blackwell GPUs and delivers better accuracy than alternative FP4 formats such as MXFP4~\citep{nvfp4_mxfp4}. NVFP4 for inference~\citep{nvpfp4_ptq_default} uses signed E2M1 values with per-block scaling over 1D blocks of size 16 along the last dimension. These per-block scales are further quantized into FP8 E4M3 using a per-tensor statically calibrated FP32 scaling factor. 


In the baseline NVFP4 PTQ recipe~\citep{nvpfp4_ptq_default}, each per-block scale is determined by the maximum absolute value in the block. 
We evaluated a range of alternative PTQ methods. The results of these experiments can be found in Appendix~\ref{sec:ptq-algorithm-ablation}.
The best overall results were obtained with a hybrid FP4 recipe: weight per-block scales were selected by minimizing weight MSE, while activation per-block scales continued to use max-based scaling. This choice is both effective and practical. Weight quantization is calibrated offline so an expensive scale search can be performed without impacting runtime performance. Activation quantization must be computed efficiently at runtime, making scale search algorithms impractical. Max-based scaling of activations provides a good trade-off between runtime performance and quantization accuracy.


In addition, we selectively promoted some layers from FP4 (W4A4) to FP8 (W8A8) or BF16 (W16A16) to further improve accuracy.
We used Model-Optimizer AutoQuantize \footnote{\url{https://github.com/NVIDIA/Model-Optimizer/tree/main/examples/llm_ptq}}, a neural architecture search (NAS) inspired method for deriving optimal mixed-precision assignments. AutoQuantize estimates per-operation sensitivity, models the performance cost of available quantization choices, and solves for the optimal layer-wise allocation using a knapsack-style optimization procedure. Its sensitivity metric follows a second-order Taylor approximation inspired by Optimal Brain Surgeon~\citep{obs}, as introduced in LLM-MQ~\citep{llmmq}.
AutoQuantize generalizes LLM-MQ beyond weight-only quantization. It supports operator-level quantization, including joint weight-and-activation quantization for GEMMs, and accounts for inference deployment constraints such as operator fusion. The detailed AutoQuantize algorithm is given in \ref{sec:autoquantize-details}.

Overall, our final NVFP4 PTQ recipe combines:
\begin{itemize}
    \item calibrated per-block weight scaling that minimizes MSE,
    \item dynamic per-block max-based activation scaling, and
    \item selective promotion of sensitive layers through Model-Optimizer AutoQuantize.
\end{itemize}
This combination addresses the accuracy loss of naive NVFP4 PTQ while preserving the runtime efficiency needed for deployment. The resulting per-operator precision assignments in the final NVFP4 checkpoint are summarized in Table~\ref{tab:precision_nvfp4}.

\begin{table}[h]
\centering
\caption{Precision settings for the backbone NVFP4 checkpoint compared with the BF16 baseline.}
\label{tab:precision_nvfp4}
\begin{tabular}{l c c c}
\toprule
\textbf{Configuration} &
\makecell{\textbf{AutoQuantize}\\\textbf{Searched?}} &
\makecell{\textbf{NVFP4}\\\textbf{Checkpoint}} &
\makecell{\textbf{BF16}\\\textbf{Baseline}} \\
\midrule
Embedding & No & BF16 & BF16 \\
\midrule
Attention QKV Projection GEMM & Yes & BF16 & BF16 \\
Attention Output Projection GEMM & Yes & FP8 / BF16 & BF16 \\
KV Cache + Attention BMM1 & No & FP8 & FP8 \\
Attention BMM2 & No & BF16 & BF16 \\
\midrule
Sparse Expert (Routed) GEMM & Yes & NVFP4 & BF16 \\
Shared Expert GEMM & Yes & NVFP4 / FP8 / BF16 & BF16 \\
MoE Latent Projection GEMM & Yes & FP8 / BF16 & BF16 \\
Router & No & FP32 & FP32 \\
\midrule
Mamba Projection GEMM & Yes & FP8 / BF16 & BF16 \\
Mamba 1D Conv & No & BF16 & BF16 \\
Mamba SSM Cache & No & FP16 & FP32 \\
\midrule
Output Layers & No & BF16 & BF16 \\
\bottomrule
\end{tabular}

\vspace{2pt}
\begin{minipage}{\linewidth}
\footnotesize
\footnotesize
For all searched backbone GEMMs, AutoQuantize considered candidate precisions from \{ \text{NVFP4}, \text{FP8}, \text{BF16} \} and selected the per-operator assignment under a quantization-sensitivity objective with an effective-precision budget of 4.75 bits. In the searched model, sparse-expert GEMMs are assigned NVFP4 throughout, attention and Mamba projection GEMMs are assigned FP8 or BF16, and shared-expert GEMMs use a mix of NVFP4, FP8, and BF16.
\end{minipage}
\end{table}

Combining AutoQuantize with the improved NVFP4 PTQ recipe produced a mostly-FP4 model, with only a small subset of layers retained in FP8 or BF16 for accuracy preservation. The full mixed-precision PTQ process completed in less than 2 hours on a single B200 node with 8 GPUs, using 512 samples from the Nemotron 3 Super SFT dataset at sequence length 4096. The resulting model achieved 99.8\% median accuracy relative to the BF16 baseline while retaining near-FP4 performance. Final evaluation results are reported in Table~\ref{tab:quant_comparison}.

\begin{table*}[h!]
\centering
\small
\setlength{\tabcolsep}{7pt}
\renewcommand{\arraystretch}{1.15}
\begin{tabular}{l|c c c}
\toprule
\textbf{Benchmark} & \textbf{N-3-Super} & \textbf{N-3-Super FP8} & \textbf{N-3-Super NVFP4} \\
\midrule

\rowcolor{black!5}
\multicolumn{4}{l}{\textbf{General Knowledge}} \\
MMLU-Pro & 83.73 & 83.63 & 83.33 \\

\midrule
\rowcolor{black!5}
\multicolumn{4}{l}{\textbf{Reasoning}} \\
HMMT Feb25 (with tools) & 94.73 & 94.38 & 95.36  \\
GPQA (no tools) & 79.23 & 79.36 & 79.42 \\
LiveCodeBench (v6 2024-08$\leftrightarrow$2025-05) & 78.69 & 78.44 & 78.44 \\
LiveCodeBench (v5 2024-07$\leftrightarrow$2024-12) & 81.19 & 80.99 & 80.56 \\
SciCode (subtask) & 42.05 & 41.38 & 40.83  \\
HLE (no tools) & 18.26 & 17.42 & 17.42  \\

\midrule
\rowcolor{black!5}
\multicolumn{4}{l}{\textbf{Agentic}} \\
Terminal Bench (hard subset) & 25.78 & 26.04 & 24.48 \\
SWE-Bench (OpenCode) & 60.47 & - & 59.90 \\
\textbf{TauBench V2} &  &  &  \\
\quad Airline & 56.25 & 56.25 & 54.75 \\
\quad Retail & 62.83 & 63.05 & 63.38 \\
\quad Telecom &64.36 & 63.93 & 63.27 \\
\quad Average & 61.15 & 61.07 & 60.46 \\

\midrule
\rowcolor{black!5}
\multicolumn{4}{l}{\textbf{Chat \& Instruction Following}} \\
IFBench (prompt) & 72.58 & 72.32 & 73.30 \\
Scale AI Multi-Challenge & 55.23 & 54.35 & 52.8 \\
Arena-Hard-V2 (Hard Prompt) & 73.88 & 76.06 & 76.00 \\

\midrule
\rowcolor{black!5}
\multicolumn{4}{l}{\textbf{Long Context}} \\
AA-LCR & 58.31 & 57.69 & 58.06 \\
RULER  128k & 97.04 & 97.17 & 96.89 \\
RULER  256k & 96.83 & 96.84 & 96.81 \\
RULER  512k & 95.22 & 95.15 & 95.21 \\
RULER  1M & 91.64 & 91.43 & 91.60 \\
\midrule
\rowcolor{black!5}
\multicolumn{4}{l}{\textbf{Multilingual}} \\
MMLU-ProX (avg over languages) & 79.35 & 79.21 & 79.37 \\

\bottomrule
\end{tabular}

\caption{
Evaluation suite for \ourmodel. We compare our \textsc{FP8} and \textsc{NVFP4} optimized models with BF16 model.
}
\label{tab:quant_comparison}
\end{table*}

\subsection{Mamba State Quantization}

In memory-bound settings, DRAM reads of the Mamba state cache (SSM cache) become a major bottleneck to decoding speed. The SSM cache is stored in FP32 by default. One option is to store the SSM cache in FP16 while arithmetic executes in FP32. In this case, the cache is fetched from memory in FP16, upcast to FP32 for the recurrent update, and then cast back to FP16 for storage. Table~\ref{tab:ssm-cache} shows experiments on an early checkpoint of Nemotron 3 Super. These results show that directly casting the SSM cache to FP16 can result in up to 40\% increase in verbosity when combined with W8A8 quantization. Even with maintaining weights and activations in BF16, casting the SSM cache to FP16 leads to up to 37\% increase in verbosity.

\begin{table}[h!]
\centering
\caption{Impact of SSM cache recipe on code benchmarks and verbosity.}
\label{tab:ssm-cache}
\resizebox{\textwidth}{!}{%
\begin{tabular}{l l c c c c c c}
\toprule
& & \multicolumn{2}{c}{\textbf{Accuracy}} & \multicolumn{2}{c}{\textbf{Completion Tokens}} & \multicolumn{2}{c}{\textbf{Verbosity Increase}} \\
\cmidrule(lr){3-4} \cmidrule(lr){5-6} \cmidrule(lr){7-8}
\textbf{Weight and} & & \textbf{livecodebench} & \textbf{scicode} & \textbf{livecodebench} & \textbf{scicode} & \textbf{livecodebench} & \textbf{scicode} \\
\textbf{Activation} & \textbf{SSM cache} & \textbf{(pass@1} & \textbf{(pass@1} & \textbf{(pass@1} & \textbf{(pass@1} & \textbf{(pass@1} & \textbf{(pass@1} \\
\textbf{Precision} & \textbf{recipe} & \textbf{avg-of-8)} & \textbf{avg-of-8)} & \textbf{avg-of-8)} & \textbf{avg-of-8)} & \textbf{avg-of-8)} & \textbf{avg-of-8)} \\
\midrule
W16A16 & FP32 & 72.91 & 40.90 & 21769 & 3680 & 0.00\% & 0.00\% \\
W16A16 & FP16 & 73.24 & 42.01 & 29812 & 3760 & 36.95\% & 2.19\% \\
W16A16 & FP16+SR (Philox 5) & 72.00 & 41.94 & 21392 & 3580 & $-$1.73\% & $-$2.72\% \\
\midrule
W8A8 & FP16 & 73.13 & 40.98 & 30536 & 3780 & 40.27\% & 2.70\% \\
W8A8 & INT16+block128 & 72.22 & 41.46 & 22406 & 3521 & 2.90\% & $-$4.30\% \\
W8A8 & FP16+SR (Philox 10) & 72.22 & 40.38 & 22120 & 3672 & 1.71\% & $-$0.74\% \\
W8A8 & FP16+SR (Philox 5) & 72.63 & 41.86 & 22159 & 3720 & 1.79\% & 1.08\% \\
W8A8 & FP16+SR (Philox 4) & 72.85 & 40.46 & 21631 & 3785 & $-$0.63\% & 2.84\% \\
W8A8 & FP16+SR (Philox 3) & 70.07 & 39.94 & 24098 & 3827 & 10.70\% & 3.98\% \\
\bottomrule
\end{tabular}%
}
\end{table}

A key challenge in quantizing the Mamba cache is that quantization error does not remain local to a single step. Because Mamba decoding is recurrent, quantization error from previous steps propagates into future steps and accumulates over time. 
This accumulation of quantization error can be seen by unrolling the recurrent update. Let the recurrent state update be \(h_t = A_t h_{t-1} + B_t x_t\), and let cache quantization at step \(t\) introduce an additive error \(e_t\), so that \(h_{q,t} = A_t h_{q,t-1} + B_t x_t + e_t\). Unrolling this recursion gives
\begin{align}
h_{q,t} &= h_t + e_t + A_t e_{t-1} + A_tA_{t-1}e_{t-2} + \cdots \nonumber\\
&\quad + A_tA_{t-1}\cdots A_2 e_1 + A_tA_{t-1}\cdots A_1 e_0 \nonumber\\
&= h_t + \sum_{i=0}^{t} \left( \prod_{j=i+1}^{t} A_j \right) e_i,
\end{align}
showing that quantization error from earlier steps is propagated through subsequent recurrent transitions and can accumulate over decoding time.

Addressing these quantization errors through changes in training (e.g., QAT or QAD) is non-trivial. Mamba training uses the chunked State Space Duality algorithm which does not explicitly materialize the recurrence relationship and the inference-time cache. Accurately modeling the recurrent decode-time cache behavior during training introduces substantial overhead. We therefore focused on training-free methods to recover the accuracy lost from cache quantization.

One way to decrease the accumulation of quantization error during PTQ is to increase the mantissa precision.
We explored this by using INT16 instead of FP16 for the SSM cache.
Naive INT16 quantization did not improve verbosity as tensor-level analysis showed that the SSM cache has a wide dynamic range. We then introduced FP32 per-block scaling over blocks of size 128 along the state dimension to increase effective dynamic range. This eliminated the verbosity issue (Table~\ref{tab:ssm-cache}).

We also explored a hypothesis that the error accumulation was tied to rounding during the cast from FP32 to FP16.
The key issue is that round to nearest, ties on even (RTNE) introduces bias in the quantization process. Because RTNE maps a given input to the same rounded value, its quantization error has zero variance but non-zero bias relative to the original value. In contrast, stochastic rounding (SR) is unbiased in expectation. In a recurrent setting, the bias from RTNE accumulates coherently over time, whereas stochastic rounding replaces this systematic drift with zero-mean noise. Based on this observation, we applied stochastic rounding before casting the cache to FP16 which fixed the verbosity issue for both the BF16 baseline and the FP8 checkpoint (Table \ref{tab:ssm-cache}).

Table \ref{tab:ssm-cache} shows accuracy and verbosity of the different SSM cache recipes for livecodebench and scicode. Both INT16 with per-block scales and FP16 with stochastic rounding were able to maintain accuracy and verbosity similar to the FP32 baseline.
We selected FP16 with stochastic rounding (SR) using Philox<5> pseudorandom number generation as the SSM cache recipe for Nemotron 3 Super for three reasons:
\begin{itemize}
    \item It does not require calculating, storing, and loading block scale factors.
    \item Blackwell provides a dedicated PTX instruction for stochastic rounding during type conversion.
    \item Blackwell supports Philox-based pseudorandom number generation through cuRAND. 
\end{itemize}
To further improve efficiency, Table \ref{tab:ssm-cache} also varies the number of Philox rounds. Increasing the number of rounds improves the statistical quality of the generated values, while reducing the number of rounds lowers pseudorandom number generation overhead. Philox<5> was chosen to maintain accuracy and verbosity while minimizing pseudorandom number generation overhead.

%% file: sections/conclusion.tex
\section{Conclusion}
\label{sec:conclusion}
We introduce \ourmodel, a 12B active and 120B total parameter MoE hybrid Mamba-Attention model with strong agentic capabilities. \ourmodel employs LatentMoE to improve accuracy and incorporates MTP layers to accelerate inference via speculative decoding. We pretrained \ourmodel on 25 trillion text tokens with low-precision NVFP4, followed by post-training on a diverse set of RL environments. Finally, we quantized the model to FP8 and NVFP4, achieving significantly higher inference throughput without sacrificing model accuracy. \ourmodel achieves up to 2.2$\times$ higher throughput than GPT-OSS-120B while maintaining higher accuracy across a wide range of tasks. We release the pre-trained, post-trained, and quantized checkpoints for \ourmodel on HuggingFace.

%% file: sections/contributors.tex
\section*{Contributors}

We thank the following people for their invaluable contributions to \ourmodelfull.

Aakshita Chandiramani, Aaron Blakeman, Abdullahi Olaoye, Abhibha Gupta, Abhilash Somasamudramath, Abhinav Khattar, Adeola Adesoba, Adi Renduchintala, Adil Asif, Aditya Agrawal, Aditya Vavre, Ahmad Kiswani, Aishwarya Padmakumar, Ajay Hotchandani, Akanksha Shukla, Akhiad Bercovich, Aleksander Ficek, Aleksandr Shaposhnikov, Alex Gronskiy, Alex Kondratenko, Alex Neefus, Alex Steiner, Alex Yang, Alexander Bukharin, Alexander Young, Ali Hatamizadeh, Ali Taghibakhshi, Alina Galiautdinova, Alisa Liu, Alok Kumar, Ameya Sunil Mahabaleshwarkar, Amir Klein, Amit Zuker, Amnon Geifman, Anahita Bhiwandiwalla, Ananth Subramaniam, Andrew Tao, Anjaney Shrivastava, Anjulie Agrusa, Ankur Srivastava, Ankur Verma, Ann Guan, Anna Shors, Annamalai Chockalingam, Anubhav Mandarwal, Aparnaa Ramani, Arham Mehta, Arti Jain, Arun Venkatesan, Asha Anoosheh, Ashwath Aithal, Ashwin Poojary, Asif Ahamed, Asit Mishra, Asli Sabanci Demiroz, Asma Kuriparambil Thekkumpate, Atefeh Sohrabizadeh, Avinash Kaur, Ayush Dattagupta, Barath Subramaniam Anandan, Bardiya Sadeghi, Barnaby Simkin, Ben Lanir, Benedikt Schifferer, Benjamin Chislett, Besmira Nushi, Bilal Kartal, Bill Thiede, Bita Darvish Rouhani, Bobby Chen, Boris Ginsburg, Brandon Norick, Branislav Kisacanin, Brian Yu, Bryan Catanzaro, Buvaneswari Mani, Carlo del Mundo, Chankyu Lee, Chanran Kim, Chantal Hwang, Chao Ni, Charles Wang, Charlie Truong, Cheng-Ping Hsieh, Chenhan Yu, Chenjie Luo, Cherie Wang, Chetan Mungekar, Chintan Patel, Chris Alexiuk, Chris Holguin, Chris Wing, Christian Munley, Christopher Parisien, Chuck Desai, Chunyang Sheng, Collin Neale, Cyril Meurillon, Dakshi Kumar, Dan Gil, Dan Su, Dane Corneil, Daniel Afrimi, Daniel Burkhardt Eliuth Triana, Daniel Egert, Daniel Fatade Douglas O'Flaherty, Daniel Lo, Daniel Rohrer, Daniel Serebrenik, Daniil Sorokin, Daria Gitman, Daria Levy, Darko Stosic, David Edelsohn, David Messina, David Mosallanezhad, David Tamok, Deena Donia, Deepak Narayanan, Devin O’Kelly, Dheeraj Peri, Dhruv Nathawani, Di Wu, Dima Rekesh, Dina Yared, Divyanshu Kakwani, Dmitry Konyagin Brandon Tuttle, Dong Ahn, Dongfu Jiang, Dorrin Poorkay, Duncan Riach, Dusan Stosic, Dustin Van Stee, Edgar Minasyan, Edward Lin, Eileen Peters Long, Elad Segal, Elena Lantz, Elena Lewis, Ellie Evans, Elliott Ning, Eric Chung, Eric Harper, Eric Pham-Hung, Eric W. Tramel, Erick Galinkin, Erik Pounds, Esti Etrog, Evan Briones, Evan Wu, Evelina Bakhturina, Evgeny Tsykunov, Ewa Dobrowolska, Farshad Saberi Movahed, Farzan Memarian, Fay Wang, Fei Jia, Felipe Soares, Felipe Vieira Frujeri, Feng Chen, Fengguang Lin, Ferenc Galko, Fortuna Zhang, Frankie Siino, Frida Hou, Gantavya Bhatt, Gargi Prasad, Geethapriya Venkataramani, Geetika Gupta, George Armstrong, Gerald Shen, Giulio Borghesi, Gordana Neskovic, Gorkem Batmaz, Grace Lam, Grace Wu, Greg Pauloski, Greyson Davis, Grigor Nalbandyan, Guoming Zhang, Guy Farber, Guyue Huang, Haifeng Qian, Haran Kumar Shiv Kumar, Harry Kim, Harsh Sharma, Hayate Iso, Hayley Ross, Herbert Hum, Herman Sahota, Hexin Wang, Himanshu Soni, Hiren Upadhyay, Huy Nguyen, Iain Cunningham, Ido Galil, Ido Shahaf, Igino Padovani, Igor Gitman, Igor Shovkun, Ikroop Dhillon, Ilya Loshchilov, Ingrid Kelly, Itamar Schen, Itay Levy, Ivan Moshkov, Izik Golan, Izzy Putterman, Jain Tu, Jan Baczek, Jan Kautz, Jane Polak Scowcroft, Janica Rosenberg, Jared Casper, Jarrod Pflum, Jason Grant, Jason Sewall, Jatin Mitra, Jeffrey Glick, Jenny Chen, Jesse Oliver, Jiacheng Xu, Jiafan Zhu, Jialin Song, Jian Zhang, Jiaqi Zeng, Jie Lou, Jill Milton, Jim Chow, Jimmy Zhang, Jinhang Choi, Jining Huang, Jocelyn Huang, Joel Caruso, Joey Conway, Joey Guman, Johan Jatko, John Kamalu, Johnny Greco, Jonathan Cohen, Jonathan Raiman, Joseph Jennings, Joyjit Daw, Juan Yu, Julio Tapia, Junkeun Yi, Jupinder Parmar, Jyothi Achar, Kari Briski, Kartik Mattoo, Katherine Cheung, Katherine Luna, Keith Wyss, Kevin Shih, Kezhi Kong, Khanh Nguyen, Khushi Bhardwaj, Kirill Buryak, Kirthi Shankar Sivamani, Konstantinos Krommydas, Kris Murphy, Krishna C. Puvvada, Krzysztof Pawelec, Kumar Anik, Laikh Tewari, Laya Sleiman, Leo Du, Leon Derczynski, Li Ding, Lilach Ilan, Lingjie Wu, Lizzie Wei, Luis Vega, Lun Su, Maarten Van Segbroeck, Maer Rodrigues de Melo, Magaret Zhang, Mahan Fathi, Makesh Narsimhan Sreedhar, Makesh Sreedhar, Makesh Tarun Chandran, Manuel Reyes Gomez, Maor Ashkenazi, Marc Cuevas, Marc Romeijn, Margaret Zhang, Mark Cai, Mark Gabel, Markus Kliegl, Martyna Patelka, Maryam Moosaei, Matthew Varacalli, Matvei Novikov, Mauricio Ferrato, Mehrzad Samadi, Melissa Corpuz, Meng Xin, Mengdi Wang, Mengru Wang, Meredith Price, Micah Schaffer, Michael Andersch, Michael Boone, Michael Evans, Michael Z Wang, Miguel Martinez, Mikail Khona, Mike Chrzanowski, Mike Hollinger, Mingyuan Ma, Minseok Lee, Mohammad Dabbah, Mohammad Shoeybi, Mostofa Patwary, Nabin Mulepati, Nader Khalil, Najeeb Nabwani, Nancy Agarwal, Nanthini Balasubramaniam, Narimane Hennouni, Narsi Kodukula, Natalie Hereth, Nathaniel Pinckney, Nave Assaf, Negar Habibi, Nestor Qin, Neta Zmora, Netanel Haber, Nick Reamaroon, Nickson Quak, Nidhi Bhatia, Nikhil Jukar, Nikki Pope, Nikolai Ludwig, Nima Tajbakhsh, Nir Ailon, Nirmal Juluru, Nirmalya De, Nowel Pitt, Oleg Rybakov, Oleksii Hrinchuk, Oleksii Kuchaiev, Olivier Delalleau, Oluwatobi Olabiyi, Omer Ullman Argov, Omri Almog, Omri Puny, Oren Tropp, Otavio Padovani, Ouye Xie, Parth Chadha, Pasha Shamis, Paul Gibbons, Pavlo Molchanov, Peter Belcak, Peter Jin, Pinky Xu, Piotr Januszewski, Pooya Jannaty, Prachi Shevate, Pradeep Thalasta, Pranav Prashant Thombre, Prasoon Varshney, Prerana Gambhir, Pritam Gundecha, Przemek Tredak, Qing Miao, Qiyu Wan, Quan Tran Minh, Rabeeh Karimi Mahabadi, Rachel Oberman, Rachit Garg, Rahul Kandu, Raina Zhong, Ran El-Yaniv, Ran Zilberstein, Rasoul Shafipour, Renee Yao, Renjie Pi, Richard Mazzarese, Richard Wang, Rick Izzo, Ridhima Singla, Rima Shahbazyan, Rishabh Garg, Ritika Borkar, Ritu Gala, Riyad Islam, Robert Clark, Robert Hesse, Roger Waleffe, Rohit Varma Kalidindi, Rohit Watve, Roi Koren, Ron Fan, Ruchika Kharwar, Ruisi Cai, Ruoxi Zhang, Russell J. Hewett, Ryan Prenger, Ryan Timbrook, Ryota Egashira, Sadegh Mahdavi, Sagar Singh Ashutosh Joshi, Sahil Modi, Samuel Kriman, Sandeep Pombra, Sanjay Kariyappa, Sanjeev Satheesh, Santiago Pombo, Saori Kaji, Satish Pasumarthi, Saurav Mishra, Saurav Muralidharan, Scott Hara, Sean Narenthiran, Sebastian Rogawski, Seonjin Na, Seonmyeong Bak, Sepehr Sameni, Seth Poulos, Shahar Mor, Shantanu Acharya, Shaona Ghosh Adam Lord, Sharath Turuvekere Sreenivas, Shaun Kotek, Shaya Gharghabi, Shelby Thomas, Sheng-Chieh Lin, Shibani Likhite, Shiqing Fan, Shiyang Chen, Shreya Gopal, Shrimai Prabhumoye, Shubham Pachori, Shubham Toshniwal, Shuo Zhang, Shuoyang Ding, Shyam Renjith, Shyamala Prayaga, Siddhartha Jain, Simeng Sun, Sirisha Rella, Sirshak Das, Smita Ithape, Sneha Harishchandra S, Somshubra Majumdar, Soumye Singhal, Sri Harsha Singudasu, Sriharsha Niverty, Stas Sergienko, Stefana Gloginic, Stefania Alborghetti, Stephen Ge, Stephen McCullough, Sugam Dipak Devare, Suguna Varshini Velury, Sukrit Rao, Sumeet Kumar Barua, Sunny Gai, Suseella Panguluri, Sushil Koundinyan, Swathi Patnam, Sweta Priyadarshi, Swetha Bhendigeri, Syeda Nahida Akter, Sylendran Arunagiri, Tailling Yuan, Talor Abramovich, Tan Bui, Tan Yu, Terry Kong, Thanh Do, Thomas Gburek, Thorgane Marques, Tiffany Moore, Tijmen Blankevoort, Tim Moon, Timothy Ma, Tiyasa Mitra, Tomasz Grzegorzek, Tomer Asida, Tomer Bar Natan, Tomer Keren, Tomer Ronen, Traian Rebedea, Trenton Starkey, Tugrul Konuk, Twinkle Vashishth, Tyler Condensa, Udi Karpas, Ushnish De, Vahid Noorozi, Vahid Noroozi, Vanshil Atul Shah, Veena Vaidyanathan, Venkat Srinivasan, Venmugil Elango, Victor Cui, Vijay Korthikanti, Vikas Mehta, Virginia Adams, Virginia Wu, Vitaly Kurin, Vitaly Lavrukhin, Vladimir Anisimov, Wan Seo, Wanli Jiang, Wasi Uddin Ahmad, Wei Du, Wei Ping, Wei-Ming Chen, Wendy Quan, Wenliang Dai, Wenwen Gao, Will Jennings, William Zhang, Xiaowei Ren, Xiaowen Xin, Xin Li, Yang Yu, Yangyi Chen, Yaniv Galron, Yashaswi Karnati, Yejin Choi, Yev Meyer, Yi-Fu Wu, Yian Zhang, Ying Lin, Yonatan Geifman, Yonggan Fu, Yoshi Suhara, Youngeun Kwon, Yuan Zhang, Yuki Huang, Zach Moshe, Zhilin Wang, Zhiyu Cheng, Zhongbo Zhu, Zhuolin Yang, Zihan Liu, Zijia Chen, Zijie Yan, Zuhair Ahmed.
\newpage

%% file: sections/appendix.tex
\appendix

\Needspace{0.8\textheight}
\section{Per-Benchmark Merge Evaluation}\label{app:per-benchmark}

\begin{figure}[H]
    \centering
    \includegraphics[width=\textwidth]{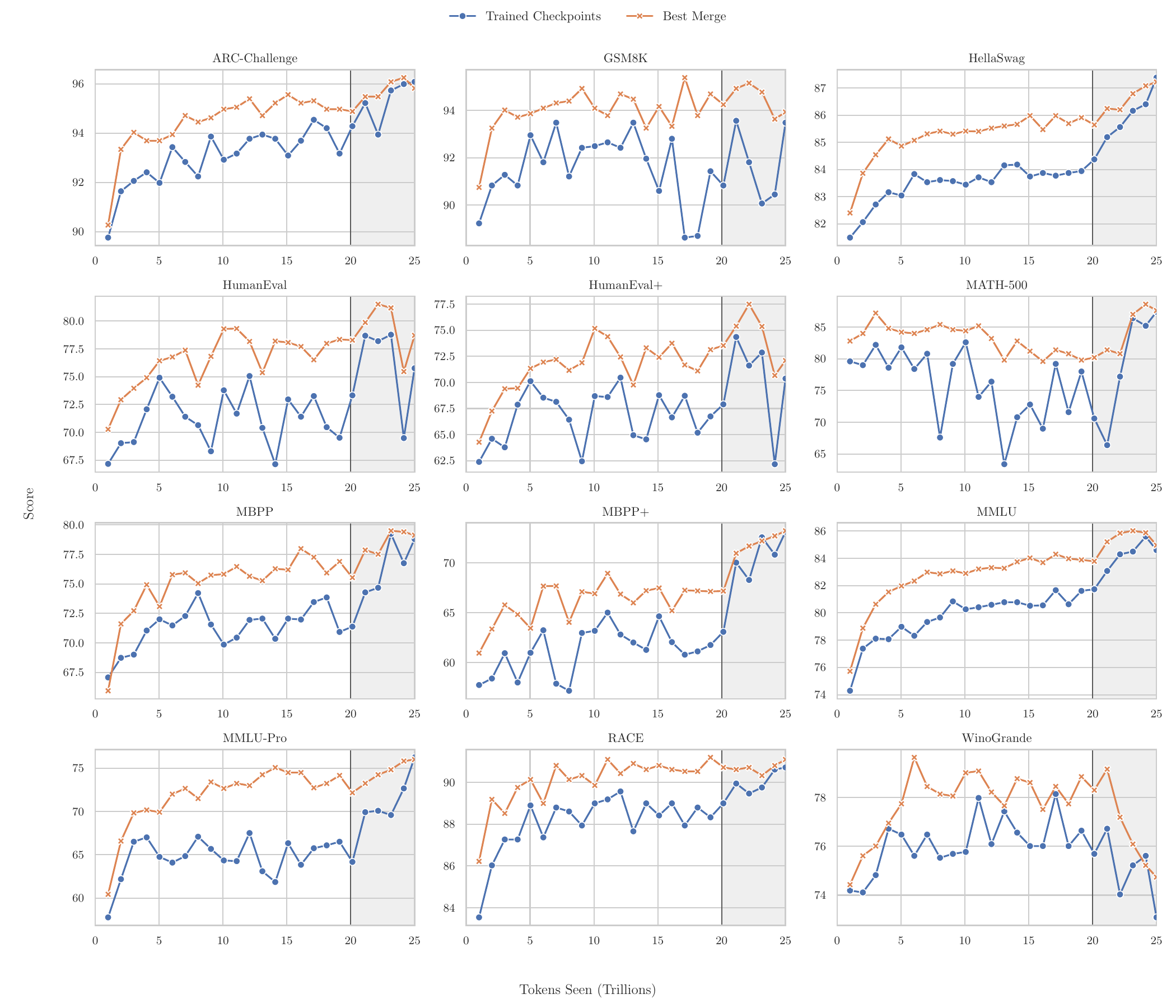}
    \caption{Per-benchmark accuracy for trained checkpoints versus the best offline checkpoint merge across the full 25T-token training run. The 12 benchmarks span general knowledge (MMLU-Pro, MMLU), code generation (HumanEval, HumanEval+, MBPP, MBPP+), mathematical reasoning (GSM8K, MATH-500), and commonsense understanding (RACE, ARC-Challenge, HellaSwag, WinoGrande). The shaded region indicates the LR decay phase.}
    \label{fig:per-benchmark}
\end{figure}

\section{FP4 Post Training Quantization (PTQ) Algorithm Details}

\subsection{PTQ Algorithm Ablation}
\label{sec:ptq-algorithm-ablation}
This section shows evaluation accuracy for various PTQ algorithms. In these experiments, all linear layers except the final classification layer and attention linear layers are quantized to NVFP4.


\begin{table}[h]
\centering
\caption{PTQ Algorithm Ablation}
\label{tab:ptq-algorithm-ablation}
\resizebox{\textwidth}{!}{%
\begin{tabular}{l p{6cm} c c c c}
\toprule
\textbf{Algorithm} & \textbf{Details} & \textbf{MMLU-Pro} & \textbf{GPQA} & \textbf{LiveCodeBench} & \textbf{AA-LCR} \\
\midrule
BF16 & --- & 83.49 & 79.92 & 72.907 & 53.00 \\
\midrule
Default NVFP4 PTQ (Baseline algorithm) & Static per-tensor scales are computed using max-value calibration; per-block scales are computed dynamically from block maximum values. & 82.99 & 79.29 & 70.18 & 55.50 \\
\midrule
Weight per-block scales minimizing MSE & Weight per-block scales are swept to minimize per-block MSE. & 83.31 & 79.92 & 71.37 & 56.75 \\
\midrule
Weight per-block scales to minimize output MSE & Weight per-block scales are swept independently to minimize GEMM output MSE. & 83.05 & 78.98 & 71.00 & 57.06 \\
\midrule
GPTQ & GPTQ~\citep{gptq} is used for weight quantization. & 83.11 & 80.05 & 69.79 & 57.87 \\
\bottomrule
\end{tabular}%
}
\end{table}

\subsection{AutoQuantize Algorithm}
\label{sec:autoquantize-details}

The sensitivity of each operation is measured at its immediate (or closest available) output using a second-order metric:
\[
S(\mathrm{Op}_i, Q_{i,f}) = \left(Y_i^{\mathrm{bf16}} - Y_i^{Q_{i,f}}\right)^\top H_i \left(Y_i^{\mathrm{bf16}} - Y_i^{Q_{i,f}}\right)
\]
where $i$ indexes operators, $Q_{i,f}$ denotes the quantized operator under format choice $f$ for operator $i$, and $H_i$ is the local Hessian approximation at the selected measurement point. The measurement point $Y_i$ can be any location where quantization error is compared against BF16; for linear layers, we use the linear-layer output.

Computing the full Hessian is expensive, so we use a diagonal Hessian approximation and estimate it empirically with the diagonal Fisher information matrix. Define
\[
\Delta Y_i = Y_i^{\mathrm{bf16}} - Y_i^{Q_{i,f}} \qquad g_i = \nabla_{Y_i} \mathcal{L}
\]
where $Y_i, g_i \in \mathbb{R}^{d}$. The resulting sensitivity proxy is
\[
S(\mathrm{Op}_i, Q_{i,f}) \approx \sum_{k=1}^{d} (\Delta Y_{i,k})^2 (g_{i,k})^2
\]

The performance cost is defined as:
\[
C(\mathrm{Op}_i, Q_{i,f}) = \mathrm{FLOPs}(\mathrm{Op}_i, Q_{i,f})
\]

AutoQuantize then solves the constrained optimization:
\[
\min_{\{f\}} \sum_i S(\mathrm{Op}_i, Q_{i,f}) \quad \mathrm{s.t.} \quad \sum_i C(\mathrm{Op}_i, Q_{i,f}) \leq B
\]
where $Q_{i,f}$ is the chosen format for operator $i$, and $B$ is the total deployment cost budget.

\subsubsection{Deployment-Restriction-Aware Search}

\paragraph{1) Linear layer fusion.}
Inference runtimes often fuse linear operators, which imposes a shared quantization format across the fused group. This fusion constraint is applied within each layer: only that layer's Q, K, and V projections are fused and required to share one quantization format. For the fused QKV projection, we model the group as one decision variable and aggregate sensitivity and cost as
\[
\begin{aligned}
S(\mathrm{Op}_{\mathrm{qkv}}, Q_{\mathrm{qkv},f}) &= S(\mathrm{Op}_{\mathrm{q}}, Q_{\mathrm{qkv},f}) + S(\mathrm{Op}_{\mathrm{k}}, Q_{\mathrm{qkv},f}) + S(\mathrm{Op}_{\mathrm{v}}, Q_{\mathrm{qkv},f}) \\
C(\mathrm{Op}_{\mathrm{qkv}}, Q_{\mathrm{qkv},f}) &= C(\mathrm{Op}_{\mathrm{q}}, Q_{\mathrm{qkv},f}) + C(\mathrm{Op}_{\mathrm{k}}, Q_{\mathrm{qkv},f}) + C(\mathrm{Op}_{\mathrm{v}}, Q_{\mathrm{qkv},f})
\end{aligned}
\]
This formulation is valid under our second-order sensitivity approximation: when the fused operator enforces one shared format but preserves additive branch contributions, the quadratic sensitivity term for the fused output remains additive across Q, K, and V.

\paragraph{2) MoE layer constraints.}
vLLM and TensorRT-LLM quantized MoE APIs require all sparse experts in a constrained MoE group to share one quantization format. This shared-format restriction is also applied within each MoE layer: only sparse experts inside the same layer are coupled. In Nemotron 3 Super, each sparse expert contains \texttt{up\_proj} and \texttt{down\_proj}, and these sparse-expert projections must therefore be assigned jointly. We formulate the sparse-expert set as one operator-level decision,
\[
\mathrm{Op}_{\mathrm{moe}} = \bigcup_{e \in \mathcal{E}} \{\mathrm{Op}_{e,\mathrm{up\_proj}}, \mathrm{Op}_{e,\mathrm{down\_proj}}\}
\]
where \(\mathrm{moe}\) denotes the MoE sparse-expert group (all coupled sparse-expert \texttt{up\_proj}/\texttt{down\_proj} operators within one MoE layer).
We measure sensitivity at the MoE block output so that the metric captures the combined contribution from all sparse experts,
\[
S(\mathrm{Op}_{\mathrm{moe}}, Q_{\mathrm{moe},f}) = \left(Y_{\mathrm{moe}}^{\mathrm{bf16}} - Y_{\mathrm{moe}}^{Q_{\mathrm{moe},f}}\right)^\top H_{\mathrm{moe}} \left(Y_{\mathrm{moe}}^{\mathrm{bf16}} - Y_{\mathrm{moe}}^{Q_{\mathrm{moe},f}}\right)
\]
and define the deployment cost as the sum over sparse experts,
\[
C(\mathrm{Op}_{\mathrm{moe}}, Q_{\mathrm{moe},f}) = \sum_{e \in \mathcal{E}} \left(C(\mathrm{Op}_{e,\mathrm{up\_proj}}, Q_{\mathrm{moe},f}) + C(\mathrm{Op}_{e,\mathrm{down\_proj}}, Q_{\mathrm{moe},f})\right)
\]
Other linear layers in the MoE block, such as latent projection layers and shared experts, are not part of this sparse-expert coupling constraint and can be assigned different quantization formats.